\documentclass[authoryear,preprint,review,12pt]{elsarticle}

\usepackage{amssymb}
\usepackage{lipsum}

\usepackage{lineno}
\usepackage{hyperref}
\usepackage{multirow}

\usepackage{mathtools}
\usepackage{abraces}
\usepackage{booktabs}

\journal{Computers and Electronics in Agriculture}

\begin{document}

\begin{frontmatter}



\title{3D Multimodal Image Registration  \\ for Plant Phenotyping}

\author[label1,label2]{Eric Stumpe}
\author[label2]{Gernot Bodner}
\author[label2]{Francesco Flagiello}
\author[label1]{Matthias Zeppelzauer}

\affiliation[label1]{organization={St. Pölten University of Applied Sciences, Institute of Creative Media Technologies},
            addressline={Campus-Platz 1},
            city={St. Pölten},
            postcode={3100},
            country={Austria}}

\affiliation[label2]{organization={University of Natural Resources and Applied Life Sciences Vienna, Department of Crop Sciences, Institute of Agronomy},
            addressline={Konrad-Lorenz-Straße 24},
            city={Tulln},
            postcode={3430},
            country={Austria}}

\begin{abstract}

The use of multiple camera technologies in a combined multimodal monitoring system for plant phenotyping offers promising benefits. Compared to configurations that only utilize a single camera technology, cross-modal patterns can be recorded that allow a more comprehensive assessment of plant phenotypes.
However, the effective utilization of cross-modal patterns is dependent on precise image registration to achieve pixel-accurate alignment, a challenge often complicated by parallax and occlusion effects inherent in plant canopy imaging. 

In this study, we propose a novel multimodal 3D image registration method that addresses these challenges by integrating depth information from a time-of-flight camera into the registration process. By leveraging depth data, our method mitigates parallax effects and thus facilitates more accurate pixel alignment across camera modalities. Additionally, we introduce an automated mechanism to identify and differentiate different types of occlusions, thereby minimizing the introduction of registration errors.

To evaluate the efficacy of our approach, we conduct experiments on a diverse image dataset comprising six distinct plant species with varying leaf geometries. Our results demonstrate the robustness of the proposed registration algorithm, showcasing its ability to achieve accurate alignment across different plant types and camera compositions. Compared to previous methods it is not reliant on detecting plant specific image features and can thereby be utilized for a wide variety of applications in plant sciences. The registration approach principally scales to arbitrary numbers of cameras with different resolutions and wavelengths. Overall, our study contributes to advancing the field of plant phenotyping by offering a robust and reliable solution for multimodal image registration. 

\end{abstract}




\begin{keyword}
plant phenotyping \sep image registration \sep multimodality \sep ray casting \sep depth camera



\end{keyword}

\end{frontmatter}




\section{Introduction}
\label{introduction}

In recent years, the application of vision systems in plant phenotyping and monitoring has garnered increasing attention and exploration. Unlike conventional sensors, these systems allow high-throughput imaging and are non-intrusive by design. Moreover, employing multiple camera technologies within a single setup has shown promise in enhancing the utility of vision systems. In such multimodal vision systems, the recorded information can be synergistically combined to identify new cross modal patterns. Compared to previous methods, this allows a deeper insight into a plant's phenotype.

In the literature for multimodal imaging in agriculture we find multiple examples showcasing that multimodal imaging yields a benefit.
\cite{gan2018photogrammetry} utilized an RGB camera to determine the maturity of citrus fruits, while simultaneously using a thermal camera to localize and distinguish the  fruits from the green tree canopy background. Similarly, other studies have successfully estimated soybean crop yields combining multispectral, thermal, and RGB imaging \cite{maimaitijiang2020soybean} or have successfully identified plants species through the combination of hyperspectral and RGB cameras \cite{salve2022multimodal}. 

However, integrating multiple camera modalities presents challenges, particularly in achieving pixel alignment due to differences in camera positions, orientations, lens parameters, and resolutions. Consequently, crucial image features such as leaf veins, leaf margins or disease patterns cannot be easily correlated or matched between the modalities. What further complicates the situation is that the different cameras may cover different spectral bands, which all have their unique characteristics (e.g., RGB and thermal cameras). This further impedes the precise alignment.

To address this challenge, various image registration methods have been proposed in the plant phenotyping literature. Many of these methods employ a purely image-based approach based on e.g., estimating a 2D affine transformation for alignment (\cite{qiu2021detection}). Such 2D registration methods cannot solve parallax and occlusion effects. As a result, registration errors arise, particularly in situations where observed plants exhibit complex geometries and close-range monitoring is performed. 

In this paper we introduce a novel registration method that overcomes the limitations associated with traditional 2D-based approaches. Our method leverages 3D information obtained from a depth camera to generate a mesh representation of the plant canopy, enabling precise pixel mapping between different cameras via ray casting. Unlike conventional approaches, our method is independent of the camera technology and the spatial camera arrangement. Furthermore, it relies solely on 3D information for registration, eliminating the need for additional computer vision algorithms tailored to specific camera types, visual pattern detection, or plant species. In addition, we introduce a classification of different occlusion and failure cases that can occur during registration and algorithms to detect and localize them automatically. This makes it possible to clearly determine for which pixels matching is possible and for which certain errors or uncertainties must be expected or where matching is not possible at all.

\begin{figure}
	\centering 
	\includegraphics[width=1\textwidth]{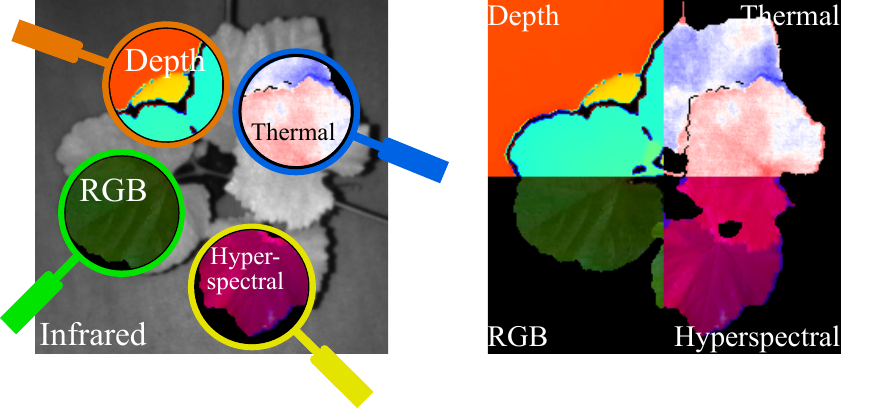}	
	\caption{Pixel alignment after registration. Our method enables to register multiple images from different types of cameras and  enables the combined processing of multimodal information at pixel-level.} 
	\label{fig_teaser}%
\end{figure}

We demonstrate the efficacy of our approach by presenting registration results for six different plant species within a camera setup comprising a hyperspectral, a thermal, and a combined RGB + infrared + depth camera (see a preview of registration results in \autoref{fig_teaser}). By showcasing the robustness and performance of our method, we aim to contribute to the advancement of plant phenotyping research, facilitating more accurate and insightful analyses across various plant species and environments.  

The concrete contributions of this paper are:
\begin{enumerate}
    \item  We present a new multimodal image registration algorithm that utilizes 3D information and ray casting to register multimodal images from different  cameras.
    \item  We develop a classification of occlusion and failure types that can occur and a method to automatically detect and classify regions that would lead to illegitimate projections, which enables us to effectively mask them.
    \item  The proposed registration method  does not only provide registered images but also registered 3D point clouds capturing geometry and measurements from different cameras.
    \item  Our method can be applied to any plant species and for any combination and number of different camera technologies (as long as one depth camera is part of the setup).
\end{enumerate}

\section{Related Work}
\label{Related Work}

\subsection{Image Registration in General}
\label{Image Registration}

Image registration, a foundational aspect of computer vision, traces its roots back to the 1970s (\cite{kuglin1975phase}). 
Typical application areas include remote sensing (\cite{fedorov2003automatic}) and applications in the medical field such as tumor growth registration over time (\cite{kyriacou1999nonlinear}).
The conventional workflow for image registration involves the identification of feature candidates (e.g., feature points) in all correlated images, followed by the matching of these candidates to establish correspondences. Subsequently, a transformation matrix (homology) is estimated, and the registration process is finalized by resampling and transforming the images using the estimated transform matrix (\cite{zitova2003image}). This approach, however, faces problems when registering images of different modalities, where the different visual characteristics impede feature matching across image modalities.  
To address this issue, \cite{maes1997multimodality} developed a method leveraging mutual information (MI) from information theory for registering computed tomography (CT) and magnetic resonance (MR) images.  
\\\\
In close-range applications, occlusion and parallax have a much stronger effect than in settings, where the camera is far away from the recorded scene (e.g., remote sensing). As a result, conventional affine or perspective transformation models fall short. Researchers have thus turned to incorporating depth information to enhance registration accuracy.  \cite{lin2019fusion} for example generated point clouds for RGB and thermal images via Shape-from-Motion (SFM) and registered both modalities in 3D space with the goal of visualizing thermal leakages in building facades.  Furthermore, \cite{schonauer20133d}  developed a head-mounted display for firefighting applications that fuses depth and thermal images to enhance visibility in low-visibility environments. On a methodological level they project the thermal information onto a 3D mesh generated from the depth information. Beyond agriculture, multimodal image registration incorporating depth information finds applications in diverse domains such as autonomous driving \cite{azam2019data}, gesture recognition \cite{van2011combining}, and industrial inspection \cite{borrmann2016spatial}, highlighting its broad utility. 

\subsection{Image Registration in Agriculture}
\label{Image Registration in Agriculture}
In agriculture, approaches to image registration range from manual methods to those leveraging depth information. \cite{cucho2020development}   registered thermal and RGB images for measuring water stress in potato crops. This was achieved by the manual selection of control points in both modalities for multiple image pairs and the subsequent computation of scale and displacement vectors.  Alternatively, \cite{liu2023yolactfusion} and \cite{clamens2021real} utilized checkerboard patterns for direct 2D registration through homography estimation. Other approaches base their registration algorithms directly on the content of the recorded plant images. \cite{yang2009automatic} first applied the Canny edge detector on the RGB and thermal images of grapevine plants and then used cross-correlation to find suitable registration parameters for translation, rotation and scale. Similarly, \cite{qiu2021detection} computed 2D image transformation matrices using local features for robust cross-modal registration. \cite{sharma2023open} and \cite{xie2023generating} both followed a two-step approach, where registration via homography estimation is computed first, followed by a fine registration step that can account for local image differences.

Even though \cite{gan2018photogrammetry}, \cite{dandrifosse2021registration} and \cite{wang2023extraction} all made use of depth cameras in their respective setups they only utilized the computation of 2D transformation models for registration and neglected the 3D information. All of the above methods utilize some form of 2D based image transformation for registration and are therefore unable to deal with occlusion and parallax effects. 
 
To address these effects, \cite{liu2018registration} utilized multiview stereo for 3D registration of RGB and multispectral images, while \cite{behmann2016generation} employed a 3D scanner for capturing point clouds of plants from multiple perspectives. A drawback of these methods is that they involve moving components, thus increasing recording time and complexity. 

A method similar to ours, introduced by  \cite{vuletic2023close}, utilized depth information from a stereo camera for point cloud registration of multispectral cameras. Specifically, they calibrated an RGB-D camera and two multispectral cameras and projected the obtained 3D points into the different camera images to yield the associated multispectral intensity values. However, using a point cloud with a given resolution they can only register as many pixels as there are 3D points. In contrast, with our ray casting approach that utilizes a 3D mesh representation, we are able to register every pixel that has a ray-mesh intersection. In addition, we integrate a built-in mechanism for the effective localization and filtering of different types of occlusion, as elaborated in this work. 

\section{Methodology}
\label{Methodology}

In the following we will first provide an overview of the entire proposed registration approach. Then we will describe each single step of the processing pipeline in separate subsections.

\subsection{Approach Overview}
\label{Approach Overview}

\begin{figure}
	\centering 
	\includegraphics[width=1\textwidth]{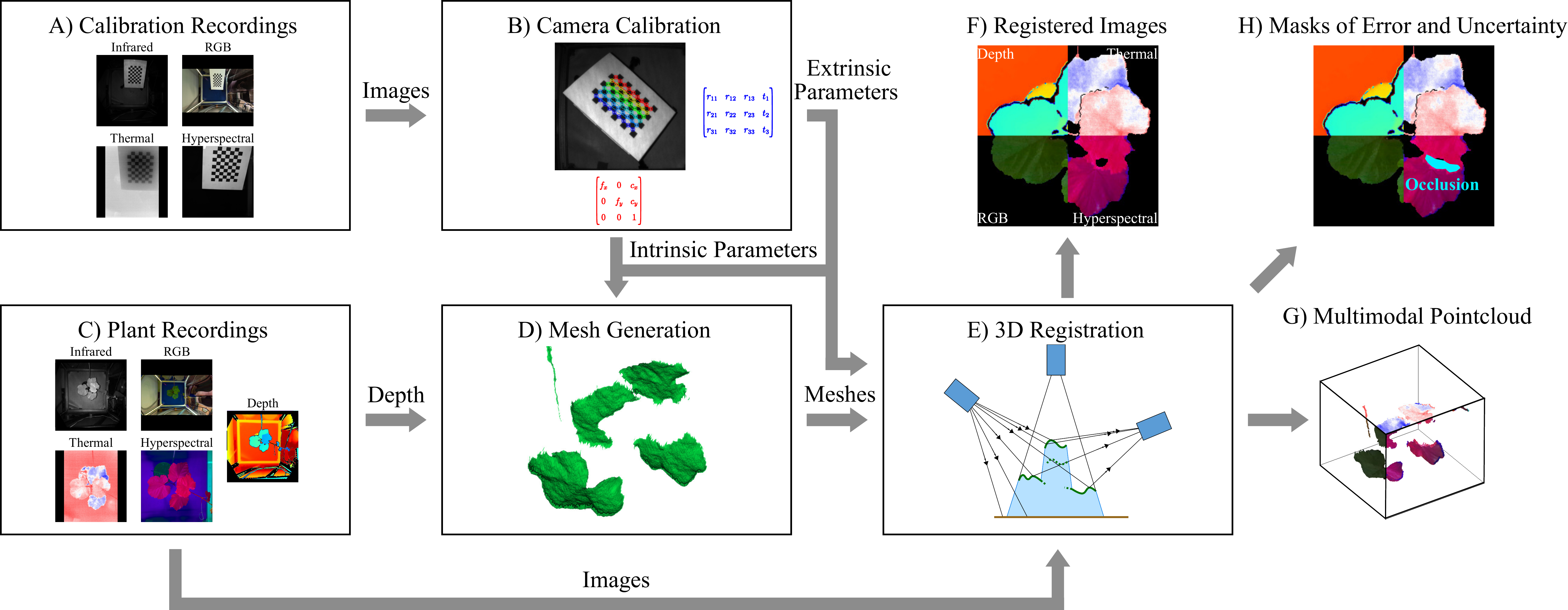}	
	\caption{Visualization of our 3D multimodal registration approach covering the entire registration pipeline.} 
	\label{fig_Methodology_overview}%
\end{figure}

\autoref{fig_Methodology_overview} provides an overview of the proposed registration approach covering the entire registration pipeline from image capturing on the input side to the registered 3D point cloud and registered images on the output side.
As a first step of our approach, all cameras that are mounted within the multimodal camera setup need to be calibrated. This is achieved by recording multiple images of a checkerboard pattern from different distances and with different orientations (A) and then using a calibration algorithm to compute the intrinsic and extrinsic parameters of each camera (B). The calibration allows us to estimate the spatial relations between the cameras in 3D. This is an important pre-requisite for ray casting, which we use to find matching pixels between the cameras. With a fully calibrated setup, plants can be positioned in the setup and imaged with each camera (C). Next, a 3D mesh of the plant canopy is created from the recorded depth map (D). The core of our registration approach is step (E), where we leverage the previously generated mesh and perform ray casting to establish pixel correspondences between the cameras. Thereafter, we perform pixel mapping to create pixel-aligned images for every camera (F). In addition, our algorithm allows us to compute a registered 3D point cloud that covers all input modalities (G).
Finally, we compute multiple masks of error and uncertainty for each image that allows us to detect, classify, and filter out invalid pixel mappings originating from different types of occlusions and  limitations introduced by the geometry between the cameras  (H). These masks can be applied to both registered images and point clouds to filter out invalid pixels and regions.

\subsection{Multimodal Camera Setup}
\label{Multimodal Camera Setup}
In our setup we use an arrangement of multiple cameras that are based on different technologies. They comprise a depth camera that can also record infrared and RGB images and both hyperspectral and thermal cameras.
The depth camera represents the centerpiece of our registration algorithm as it enables us to map pixel information between cameras.
A more detailed explanation of the hardware setup is provided in \autoref{Hardware Setup}. Within this paper we will refer to the depth camera as $C_D$. Since our approach works for all kinds of camera configurations we keep the notation of all cameras general. Of all remaining N non-depth camera we can choose a target view camera $C_T$. This means that our goal is to align all pixels of all other cameras to the image space of  $C_T$. A camera that is used to extract pixel values from, we call source view camera $C_S$.

\subsection{Fundamentals: 3D Projections }
\label{Projections in 3D space}

\begin{figure}
	\centering 
	\includegraphics[width=0.7\textwidth]{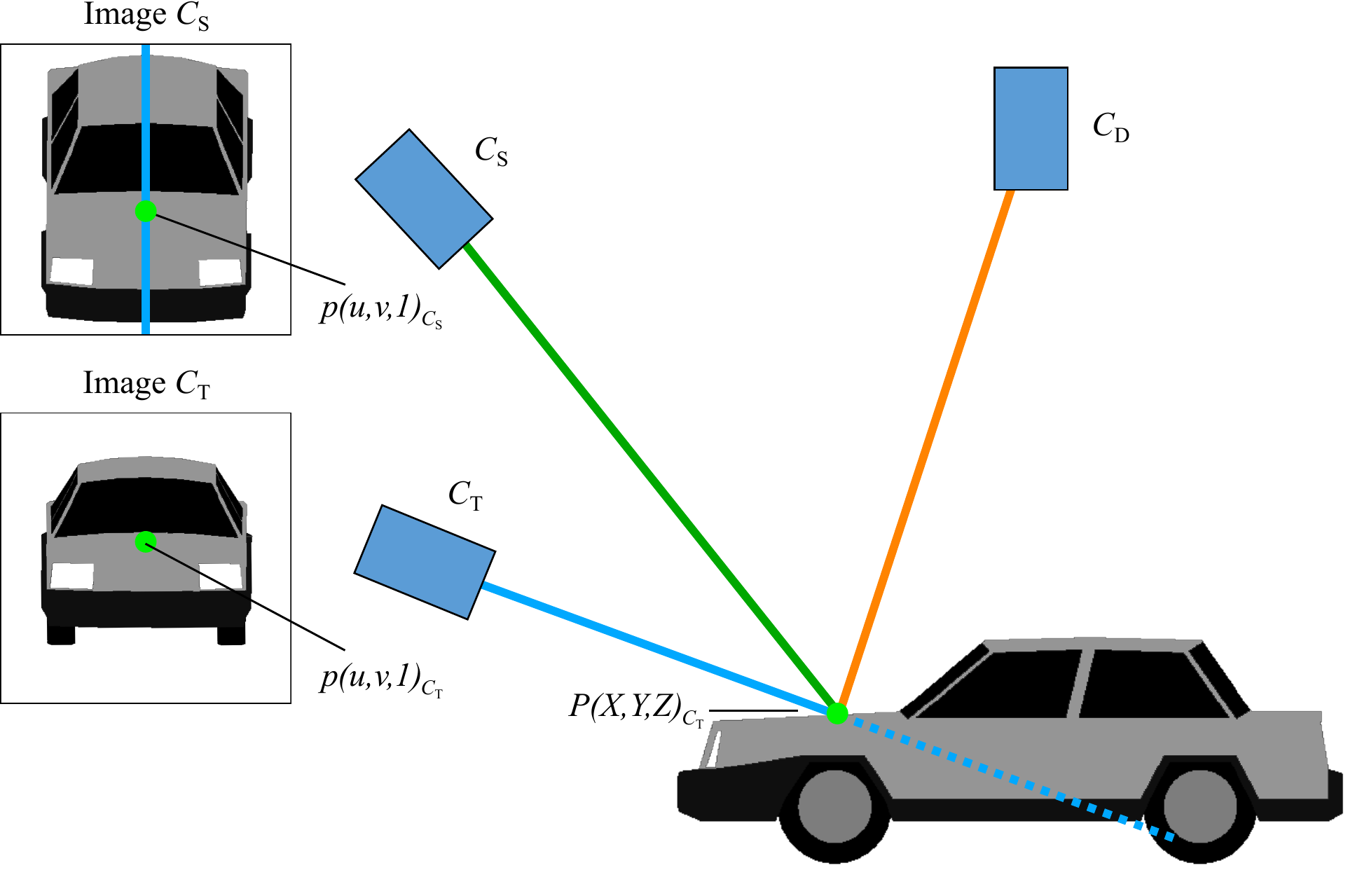}	
	\caption{3D projection principle. Geometrical example for projecting pixel information for registration. $C_D$ is the depth camera, while $C_T$ and $C_S$ are arbitrary cameras at different positions that represent the target view and source view respectively. Image $C_T$ and Image $C_S$ show the images recorded by both cameras of the observed car in 3D space (bottom right). The blue line represents the epipolar line that corresponds to $p(u,v,1)_{C_T}$ being projected into 3D space. The green and orange lines represent the projection lines of $C_T$ and $C_D$ respectively.} 
	\label{fig_car_projection}%
\end{figure}

Before continuing with the explanations of the different steps in our approach we give an introduction to the underlying projection principle that is the basis for our algorithm. To facilitate a better comprehension of the principles to the reader we provide a graphical explanation.  In \autoref{fig_car_projection}, we demonstrate the concept for a car and three cameras $C_D$, $C_T$ and $C_S$.  

For each camera $C$, we can describe with the pinhole model how arbitrary 3D points  $P(X,Y,Z)_C$ of the observed scene  can be projected into the camera’s image space $p(u, v, 1)_C$. In \autoref{fig_car_projection}, $P(X,Y,Z)_{C_T}$, which is a 3D point (green) on the car in 3D space can be projected into image of $C_T$, where it than located at the pixel $p(u, v, 1)_{C_T}$ (green). This projection can be accomplished with the formula:

\begin{equation}
s \begin{bmatrix} u \\ v \\ 1 \end{bmatrix}_{C} = 
\underbrace{\begin{aligned}
\begin{bmatrix}
f_{x} & 0 & c_{x} \\
0 & f_{y} & c_{y} \\
0 & 0 & 1 
\end{bmatrix}\end{aligned}
}_{K_{c}} \begin{bmatrix} X \\Y\\ Z \end{bmatrix}_{C}.
\label{eq:1}
\end{equation}

To enable the projection, we need to apply the intrinsic camera matrix $K_C$ that can be obtained via camera calibration. In this equation, $f_x$ and $f_y$ represent the focal lengths and $c_x$ and $c_y$ represent the principal points for both image axes. Additionally, the scale factor, denoted as $s$, equals $Z$ in this context. Geometrically, the focal length of the intrinsic camera matrix $K_C$ will influence at which angle light rays (blue line in \autoref{fig_car_projection}) originating from a point $P(X,Y,Z)_{C_T}$ will arrive in the lens of the camera $C_T$.

If we want to do the reverse process and find out which 3D point in the scene belongs to a given image coordinate, we can invert \autoref{eq:1} to get

\begin{equation}
\frac{1}{s} \begin{bmatrix} x" \\ y" \\ 1 \end{bmatrix}_{C} = 
\underbrace{\begin{aligned}
\begin{bmatrix}
f_{x}^-1 & 0 & c_{x} \cdot f_{x}^-1 \\
0 & f_{y}^-1 & c_{y} \cdot f_{y}^-1\\
0 & 0 & 1 
\end{bmatrix}\end{aligned}
}_{K_{C}^-1} \begin{bmatrix} u \\ v \\ 1 \end{bmatrix}_{C}.
\label{eq:2}
\end{equation}

However, in this order the scale is unknown and instead we only get a 3D epipolar line $(x”, y”, 1)\frac{1}{s}$. This 3D epipolar line is visualized in blue in \autoref{fig_car_projection}. Without any additional information $p(u,v,1)_{C_T}$ could originate from any 3D point along the epipolar line.

This is where auxiliary 3D information from the depth camera $C_D$ within the setup comes into play since it allows us to resolve $s$ by providing the matching depth information and hence get the corresponding $P(X, Y, Z)_{C_T}$ coordinates (orange line in \autoref{fig_car_projection}). For projections between 2D and 3D we need to consider that $(X, Y, Z)_C$ must be based in the associated camera coordinate system in order to compute the projections. To transform the 3D coordinates from one camera coordinate system $C_T$ to another $C_S$, we can use  

\begin{equation}
\begin{bmatrix} X \\Y\\ Z \\ 1 \end{bmatrix}_{C_S} = 
\underbrace{\begin{aligned}
\begin{bmatrix}
r_{11} & r_{12} & r_{13} & t_{1} \\
r_{21} & r_{22} & r_{23} & t_{2} \\
r_{31} & r_{32} & r_{33} & t_{3} \\
\end{bmatrix}\end{aligned}
}_{M_{TS}} \begin{bmatrix} X \\Y\\ Z \\ 1 \end{bmatrix}_{C_T}.
\label{eq:3}
\end{equation}

The extrinsic camera matrix $M_{
TS}$ consists of a rotational term $R$ ($r_{11},...,r_{33}$) and a translational term $T$ ($t_1,t_2,t_3$). 

Having introduced these equations, we can use them in our car example.
If we want to find out which pixel $p(u,v,1)_{C_S}$ corresponds to pixel $p(u,v,1)_{C_T}$ in the target view image we pursue the following steps. For $p(u,v,1)_{C_T}$, we compute the corresponding epipolar line $(x”, y”, 1)_{C_T}\frac{1}{s}$ (blue line) via \autoref{eq:2}. With the help of depth information we can determine the corresponding 3D point $P(X,Y,Z)_{C_T}$ (details in \autoref{Registration}). 
We then convert $P(X,Y,Z)_{C_T}$ to $P(X,Y,Z)_{C_S}$ with \autoref{eq:3} and use \autoref{eq:2} to project the 3D point into image $C_S$ to yield $p(u,v,1)_{C_S}$. Note that without any depth information, we would only be able to project the complete 3D epipolar line into the image of $C_S$ to compute a 2D line that represents all potential locations of $p(u,v,1)_{C_S}$ (blue line in Image $C_S$).

\subsection{Camera Calibration (A, B)}
\label{Camera Calibration}

Our registration method is based on transporting pixel information via ray casting between different cameras. A prerequisite for our algorithm is that the intrinsic and extrinsic camera parameters $K_C$ and $M_{TS}$ as well as lens distortion coefficients are known for each camera in the setup. 


In our approach we use the camera calibration routine from OpenCV based on the calibration algorithm proposed by \cite{zhang2000flexible}. A calibration checkerboard with known geometry is recorded by each camera in the setup. This process is repeated for a series of different 3D checkerboard positions and orientations after which we are able to compute both intrinsic and extrinsic camera parameters.

Essential to our registration algorithm is to establish the spatial relationship between the depth camera and all other cameras (to enable ray casting). The depth image (in our setup recorded by a Time-of-Flight (ToF) camera), however, cannot capture the tiles of the flat checkerboard calibration target. It is thus essential to employ a depth camera that also captures an image (RGB or infrared) next to the depth map and which provides intrinsic and extrinsic parameters for the different camera modalities). In our setup, the ToF camera produces precisely aligned accompanying infrared images. Thus, the extrinsic parameters for the depth camera can be estimated via the accompanying infrared, which captures the checkerboard target well. For the thermal camera we heated up the checker board tiles to utilize the difference in heat absorption of black and white checker tiles to make the checkerboard visible (see \autoref{Calibration} for details)

\subsection{Plant Recordings (C)}
\label{Plant Recordings}

After the camera setup is calibrated, we capture images of the plant specimen. To this end, the plant is placed below the cameras such that it is visible to all cameras in the setup. Note that the overlap in view ports of the different cameras needs to be optimized/maximized before camera calibration. Further note that the camera calibration is only precise for a limited depth range, i.e., the depth range covered during camera calibration. The plant should thus be positioned in a way that it best covers the calibrated depth range.  After taking an image with each camera the processing stage can be started.

\subsection{Mesh Generation}
\label{Mesh Generation}

\begin{figure}
	\centering 
	\includegraphics[width=0.5\textwidth]{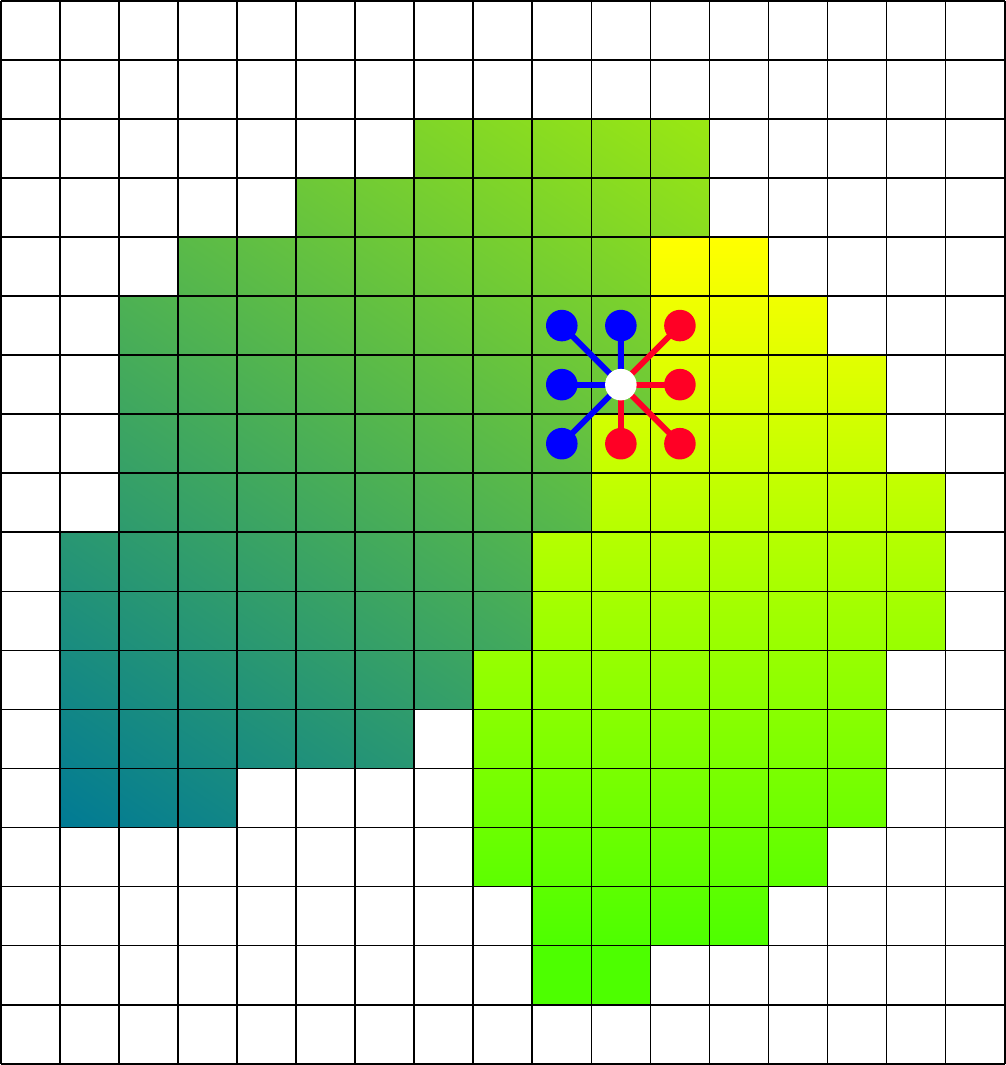}	
	\caption{Schematic visualization of the generation of the 3D mesh from a depthmap for the example of two adjacent leaves. The depthmap distance of each pixel is represented by a color gradient of blue (near) to yellow (far)
 Each pixel (white dot for example) is converted to a mesh vertex and connected with its 8 neighbors via mesh edges. Edges are omitted if the vertical angle is over 15°, which can occur for pixels at adjacent leaves (red lines). } 
	\label{fig_mesh_drawing}%
\end{figure}

As detailed in the previous section we need to include 3D information to solve for \autoref{eq:2}. This means for each pixel in an image we need to know the associated depth.
One option to accomplish this would be to work with 3D point cloud representations. With those we could cast rays from the target view into the scene and  compute the closest 3D points per ray to determine the depth associated with each pixel. Alternatively, we could use the 3D points directly for transporting  pixel information between different cameras, i.e., we would start with a 3D point, project it into the two camera views, save both pixel positions and then transfer the pixel from the source view to the target view as done by \cite{vuletic2023close}. This way we are however completely dependent on the number of 3D points available i.e., the spatial coverage within the recorded point cloud. With a very sparse point cloud we could only partially register pixels of high-resolution images.
There is also another disadvantage shared by the two latter techniques. Since 3D point clouds are sparse data structures, a projected ray can pass through the free space between 3D points of one object (e.g., a leaf) and be the closest to a 3D point of an underlying structure (e.g., a leaf below). This makes it difficult to detect the occurance of different types of occlusion.\\\\
We propose a more efficient approach by using a 3D triangle mesh instead of a 3D point cloud.
This way we can cast rays from each pixel and record the intersection points with the mesh.
By utilizing the Python package Open3D that uses functions from the efficient ray casting library Intel Embree we can therefore achieve better computation speed. Within a mesh of a complex object, partial elements (e.g., leaves) are also impermeable to cast rays, which simplifies the detection of different projection errors (see \autoref{Projection Cases}).
Finally, using meshes allows us to work with arbitrary image resolutions. \\\\
To generate the 3D mesh of the observed object we call object mesh $\mathcal{M}_o$, we use the recorded depth map from the depth camera in the setup. In our experiments $\mathcal{M}_o$ represents a plant canopy, but our method also works for any objects of complex geometry.
As a first step, we define a 3D Region of Interest (ROI) to filter out depth information that is outside of the plant canopy. The ROI can be selected by considering the distance of the depth camera to the ground (Z) and the expected canopy extent (X, Y). Usually this ROI only has to be defined once for a specific camera setup. Next, we compute the corresponding 3D points for each depth pixel via \autoref{eq:2} and designate them as the vertices of the canopy triangle mesh. The edges of the mesh are constructed by considering the immediate 8-pixel neighborhoods of each depth pixel. Therefore, at maximum a vertex can be connected to 8 edges and be part of 4 neighboring mesh triangles.
Additionally, in the mesh we want to prevent that vertices at the border of leaves are connected to those of adjacent leaves. To filter out affected mesh edges a maximum vertical angle of 15° is set between adjacent vertices.
A visualization of the mesh generation process from the depthmap is provided in \autoref{fig_mesh_drawing}. In the figure, two schematic overlapping leaves in a depthmap are shown. The depth is visualized by a color gradient from blue to yellow representing an increasing distance to the depth camera.
The white dot represents a sample depth pixel that is converted to a mesh vertex and connected via edges to its neighboring vertices. However, connections with a large vertical angle (red) are excluded.


\subsection{3D Registration (E)}
\label{Registration}

With the completion of the mesh generation process we can proceed to register the pixels of one camera target view image $(u,v,1)_{C_T}$ with those of another source view camera $(u,v,1)_{C_S}$. With our method, any of the cameras can be chosen as the target view or source view.
Subsequently, for every pixel coordinate $(u,v,1)_{C_T}$ of the target view, we calculate the corresponding 3D epipolar line $(x”,y”,1)\frac{1}{s}$ from \autoref{eq:2}. These epipolar lines are then cast as rays into $\mathcal{M}_o$ to compute 3D intersection points $(X,Y,Z)_{C_T}$, effectively solving the 2D to 3D correspondence problem. After transforming these points to $(X,Y,Z)_{C_S}$ via \autoref{eq:3}, we project them into the image space of $C_S$ using \autoref{eq:1} to get $(u,v,1)_{C_S}$. Consequently, for each pixel $(u,v,1)_{C_T}$ of $C_T$ we obtain the matching pixel coordinates $(u,v,1)_{C_S}$ of $C_S$ and their respective image pixel values to construct the registered image through mapping (see next Section for details). 

\subsection{Registered Images (F)}
\label{Registered Images}

To create a registered image for all available modalities, we select a target view $C_T$ = $C_i$ (e.g., the infrared image) and perform the above routine over all other of the remaining $n-1$ camera images $C_S$ = $C_j$ with $j=1,..,n$ and $j \neq i$. This process can be performed for any image modality as the selected target view.

\subsection{Multimodal Point Clouds (G)}
\label{Multimodal Point Clouds}

To create a multimodal point cloud, we can almost follow the same approach.
Again we select a target view $C_T$. However, here we do not map the pixel values from the originating views $C_S$ back to the target image. Instead we use all recorded 3D intersection points of rays with $\mathcal{M}_o$ to create a 3D point cloud and map the image pixel values back to the intersection points.

\subsection{Classification of Projection Cases (H)}
\label{Projection Cases}

The registration process described so far follows the assumption that nothing interferes with the path of a ray from one modality to another. In reality, plant canopies are characterized by the complex geometry of overlapping and discontinuous surfaces and thus the above assumption is not valid in general. As a result, different types of errors and cases of uncertainties may appear. In this subsection we introduce a classification of the different cases of error and uncertainty -- short: $P_i$ -- that may occur and provide a solution for their automated identification and localization during the registration process in \autoref{Projection Error Detection}. An overview of all projection cases is provided in \autoref{tab:certainties}.

\begin{figure}
	\centering 
	\includegraphics[width=1\textwidth]{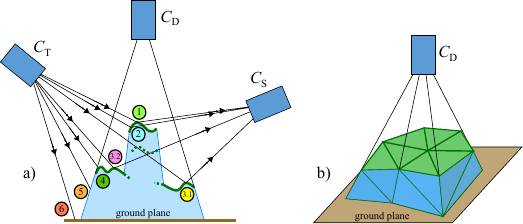}	
	\caption{Projection Cases. \textbf{a)} Different cases of projection errors and uncertainties that can occur between different cameras $C_T$ and $C_S$, with $C_D$ being the depth camera: 1) legitimate projections, 2) occlusion error, 3.1) incoming uncertain correspondence, 3.2) uncertain correspondence.
    The cases 4, 5 and 6 are only relevant from the perspective of the target camera $C_T$.
    4) indicates the certain canopy area. 5) represents the uncertain area, while 6) represents the certain background area. 
    \textbf{b)} the uncertainty mesh: The green polygons exemplify a simplified canopy object mesh $\mathcal{M}_o$. The brown area represents the ground plane. By shooting rays from the depth camera $C_D$ through the vertices of border edges of $\mathcal{M}_o$ we find the intersections with the ground plane, which allows us to construct the uncertainty mesh $\mathcal{M}_u$ (blue mesh).} 
	\label{fig_projections}%
\end{figure}

To illustrate the different cases $P_i$ that may occur, \autoref{fig_projections} (a) shows a schematic projection scene featuring three different cameras $C_D$, $C_T$, $C_S$ and the intersecting view of the object mesh $\mathcal{M}_o$, which is a plant canopy with leaves depicted in green. Notably, $C_D$ denotes the depth camera, indicating that $\mathcal{M}_o$ originates from the depth image captured by $C_D$. Some of the leaves of the canopy are not visible from $C_D$ because they are occluded by other leaves. These occluded object parts are visualised via dashed lines. $C_T$ and $C_S$ are further cameras of the multimodal setup of arbitrary modality. The presented schematic is depicted for a plant canopy, but the derived projection cases would also apply for other complex objects.

We can differentiate the following cases $P_1$ to $P_6$ that may occur during 3D registration in the proposed camera setup.
\paragraph{$P_1$: Legitimate Projection} 
Case $P_1$ displays a clearly solvable matching case. The ray is cast from target view $C_T$, hits $\mathcal{M}_o$ and is projected into $C_S$  to get the corresponding pixel value for registration. There are no obstacles in between.

\paragraph{$P_2$: Occlusion} 
Case $P_2$ demonstrates a typical occlusion scenario known from stereo vision. Once again, a ray originating from target view $C_T$ intersects with $\mathcal{M}_o$. However, en route to camera $C_S$, the outgoing ray intersects with another part of the canopy. Given that the latter intersection point is closer to the camera origin of $C_S$, only this particular part of the canopy will be visible to $C_S$. Consequently, without checking for this type of error, a mismatched pixel value would be incorrectly mapped back to $C_T$.  

\paragraph{$P_{3.1}$, $P_{3.2}$: Uncertain Correspondence} 
The generated $\mathcal{M}_o$ only represents 3D information of the object mesh surface that is visible to the depth camera $C_D$ (depicted by solid green lines).  However, given the multi-layered nature of plant canopies, it is plausible that concealed leaves, unseen by camera $C_D$, exist (illustrated by dashed green lines in \autoref{fig_projections} a)). This obscured region is visualized in blue. Two distinct cases of uncertainty emerge from this occluded space: 
In case $P_{3.1}$, the incoming ray first intersects the occluded space before reaching the mesh surface. 
Consequently, there exists a risk that the computed 3D intersection point might not be the correct match for the pixel coordinate $(u,v,1)_{C_T}$. Instead, the pixel value in the image of $C_T$ may originate from an obscured leaf situated in front of the intersection point. 
Contrarily, in $P_{3.2}$, the outgoing ray intersects the occluded space. Although the resulting 3D intersection point is valid, uncertainty exists regarding the accuracy of the intersection point  obtained from the ray going towards $C_S$. This uncertainty arises from the possibility of an occluded leaf positioned closer to the coordinate origin of $C_S$. 

\paragraph{$P_4$: Certain Object Area}
Independently from computing mappings between cameras we are able to make assessments about the content of different regions within each image modality. This can be beneficial prior information for further plant analysis algorithms such as leaf segmentation to enable masking out regions where no cross-modal matching can be performed. Case $P_4$ represents the area  where incoming rays first hit $\mathcal{M}_o$ and we can be certain that the associated pixels belong to the object.

\paragraph{$P_5$: Uncertain Object Area} 
Additionally to the case above, there are areas for which we cannot determine whether the associated rays will intersect with parts of the object. For every originating ray that first hits the occluded space (blue in \autoref{fig_projections}), the pixel can either belong to the background or the object itself since this part is not visible to the depth camera.

\paragraph{$P_6$: Certain Background Area}
Finally, every ray that neither hits the object mesh $\mathcal{M}_o$ nor the occluded space can confidently be attributed be part of the background and not the object itself.

\begin{table}[ht]
\centering
\resizebox{1\columnwidth}{!}{
\begin{tabular}{l|l|cc|cc}
\textbf{Case} & \textbf{Certainty} & \multicolumn{2}{c|}{\textbf{Cameras Involved}} & \multicolumn{2}{c}{\textbf{Mesh Intersections}} \\
              &                         & \textbf{$C_T$} & \textbf{$C_S$} & \textbf{$\mathcal{M}_o$} & \textbf{$\mathcal{M}_u$} \\
\midrule
$P_1$   & Certain Match          &$\checkmark$&$\checkmark$&$\checkmark$&   \\
$P_2$   & Certain False Match    &$\checkmark$&$\checkmark$&$\checkmark$&   \\
$P_{3.1}$ & Uncertain Match        &$\checkmark$&$\checkmark$&$\checkmark$&$\checkmark$\\
$P_{3.2}$ & Uncertain Match        &$\checkmark$&$\checkmark$&$\checkmark$&$\checkmark$\\
$P_4$   & Certain Object Area    &$\checkmark$&   &$\checkmark$&   \\
$P_5$   & Uncertain Object Area  &$\checkmark$&   &   &$\checkmark$\\
$P_6$   & Certain Background Area&$\checkmark$&   &   &   \\
\bottomrule
\end{tabular}
}
\caption{Overview of the projection cases. The cameras involved column describes which cameras target view $C_T$ or source view $C_S$ are needed to identify the projection case. To differentiate between different projection cases we also need to determine whether just the object mesh $\mathcal{M}_o$ and/or the uncertainty mesh $\mathcal{M}_u$ intersect with the incoming and outgoing ray.}
\label{tab:certainties}
\end{table}

\subsection{Detection of Projection Cases (H)}
\label{Projection Error Detection}

In this section we present how we identify the different projection cases of error and certainty  algorithmically. We illustrate the procedures at the example of two cameras $C_T$ and $C_S$ as also illustrated in \autoref{fig_projections}a).

\paragraph{$P_1$: Legitimate Projection} 
To determine whether a projection is legitimate we check if the incoming ray from $C_T$ first intersects with $\mathcal{M}_o$ and not the occluded space. If  the other projection cases $P_2$-$P_{3.2}$ do not apply, we can assert that the projection is valid.

\paragraph{$P_2$: Occlusion}
Following the computation of the 3D intersection point of the incoming ray from $C_T$ and identifying the potential pixel position in $C_S$, we cast another ray from $C_S$ towards the 3D intersection point. By confirming if this ray intersects with $\mathcal{M}_o$ at a closer distance, we can ascertain the occurrence of an occlusion. 

\paragraph{$P_{3.1}$, $P_{3.2}$: Uncertain Correspondence}
We propose a technique to identify projection cases $P_{3.1}$ and $P_{3.2}$ by representing the occluded space as a separate mesh entity. This approach is schematically illustrated in \autoref{fig_projections} b). Initially, we take the canopy object mesh $\mathcal{M}_o$ (depicted as green triangles) to isolate edges that delineate leaf boundaries. These edges are distinguishable as they belong to only one triangle, unlike other edges in the mesh that are part of two triangles. Subsequently, we project the vertices of these boundary edges onto the ground plane according to the cameras intrinsic parameters via \autoref{eq:1}. The ground plane is the $z$ plane of the previously defined Region of Interest (RoI). Then we record the resulting intersection points on the ground. Utilizing these boundary edge vertices and ground plane intersection points, we construct a new mesh segment to represent the occluded space (depicted in light blue) that we refer to as uncertainty mesh $\mathcal{M}_u$. 

This approach enables the identification of both types of uncertain correspondence cases. For case $P_{3.1}$, the presence of this error is determined by assessing whether incoming rays intersect $\mathcal{M}_u$ prior to reaching $\mathcal{M}_o$. Conversely, for case $P_{3.2}$, akin to the simple occlusions ($P_{2}$), we cast a ray from camera $C_S$ and examine whether it intersects with $\mathcal{M}_u$ before reaching the intersection point with $\mathcal{M}_o$.

\paragraph{$P_4$, $P_5$, $P_6$: Certainty Areas}
The three cases $P_4$ to $P_6$  can be  identified by computing the first intersections of the incoming ray of $C_T$. If the ray hits $\mathcal{M}_o$ first, it can be attributed to the \textit{certain canopy area} (case $P_4$) and can subsequently be assigned to one of the cases $P_1$, $P_2$, or $P_{3.2}$. If the incoming ray instead hits $\mathcal{M}_u$ first, the pixel belongs to the \textit{uncertain object area} (case $P_5$). If the ray hits neither of both meshes $\mathcal{M}_o$ and $\mathcal{M}_u$ we assign it the \textit{certain background area} (case $P_6$).  

\subsection{Projections for N cameras}
So far we have presented our methodology for computing registrations between only two cameras  with the assistance of depth information. Now we outline our procedure for computing registrations for N cameras. First, we select our target view $C_T$ = $C_i$ into which we would like to project all other views $C_S$ = $C_j$ , where $j \in \{1, 2, \ldots, N\}$ and $j \neq i$.
As a first step we use all pixels of $C_T$ and compute all incoming rays. We then record all intersections with $\mathcal{M}_o$ and $\mathcal{M}_u$. We only have to do this step once since the incoming rays and intersection points are the same for all pairings. Next, for all $N-1$ source views $C_j$ we compute the outgoing rays from the recorded intersection points with $\mathcal{M}_o$ and determine the matching pixel positions. Consequently, we also perform all checks to determine all projection cases $P_1$ - $P_6$, for each combination of $C_i$ and $C_j$ based on the computed incoming and outgoing rays.





\section{Experimental Setup}
\label{Experimental Setup}

We investigate the capabilities of the proposed cross-modal 3D image registration approach in several experiments with different plant species using a custom-built image capturing setup. Here we present the setup of the experiments, all necessary pre-processing steps, the dataset  and the evaluation metrics that allow us to measure the registration accuracy.

\subsection{Hardware Setup}
\label{Hardware Setup}

\begin{figure} [ht]
	\centering 
	\includegraphics[width=1\textwidth]{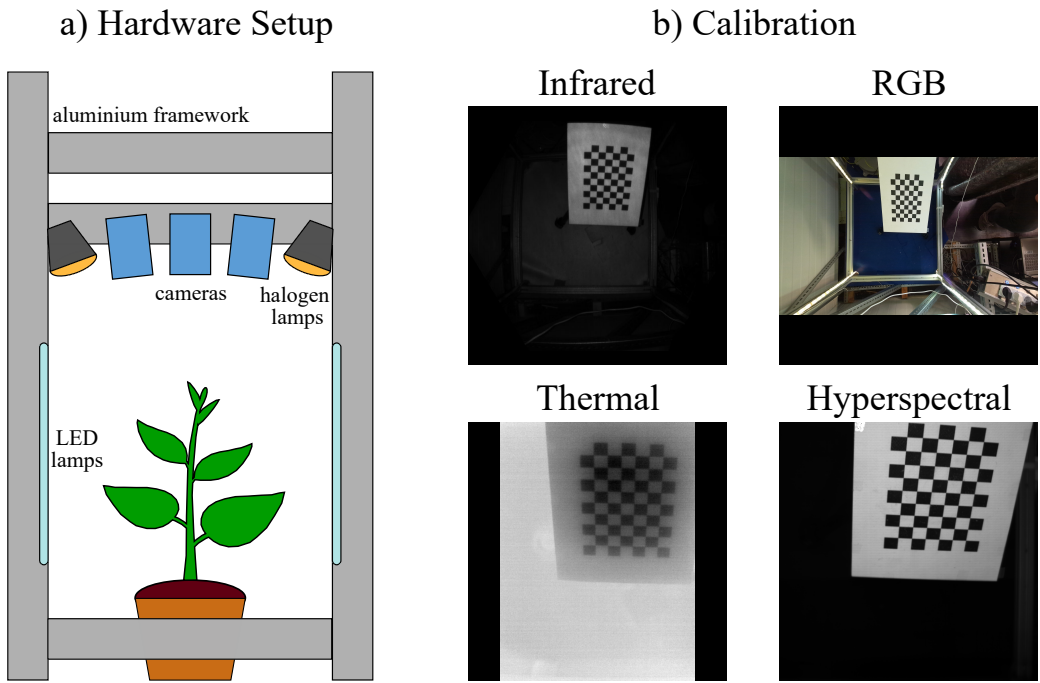}	
	\caption{a) Overview of the multimodal camera setup. The three cameras (depth + RGB + infrared), thermal, and hyperspectral) are visualized in blue and are mounted at the top of the aluminium profile construction (gray). In addition we have two sources of light: halogen lamps (yellow) and LED strips (light blue). b) Camera calibration: During calibration we move a checker board within the scene and record it with every camera. After the calibration we get the intrinsic and extrinsic camera parameters.} 
	\label{fig_Overview_setup}%
\end{figure}

Our experimental setup comprises a multimodal camera system featuring three distinct cameras, each tailored to capture different aspects of the scene, namely a combined  RGB, infrared and depth camera, a thermal camera and a hyperspectral camera. \autoref{fig_Overview_setup} a) displays the overall aluminium profile framework (gray) that holds all the cameras and the attached lights. The cameras (blue) are mounted at the top of the framework.

\paragraph{RGBD Camera} For recording RGB, depth, and infrared imagery, we employ the Azure Kinect camera by Microsoft, leveraging its time-of-flight (ToF) technology. The RGB images are captured at a high resolution of 3840 x 2160 pixels, offering a field of view of 90° × 59°. Simultaneously, the depth and infrared images are acquired at a resolution of 640 x 720 pixels, with a slightly narrower field of view measuring 75° × 65°. The depth images have an accuracy of $<$11mm + 0.1\% of the recorded distance. In our setup we  record plant canopies within a depth range of 30 cm and 120 cm (ground plane) away from the depth camera.

\paragraph{Thermal Camera} Thermal imaging is facilitated by the VarioCAM HDx Head 620 S camera from Infratec(R). This camera delivers thermal images at a resolution of 640 x 480 pixels, offering a field of view measuring 32.7° × 24°. The manufacturer specifies a temperature measurement accuracy of ±2\%. 

\paragraph{Hyperspectral Camera} Our setup incorporates the Specim IQ hyperspectral camera by Specim. Operating on a linescan principle, this camera captures images at a resolution of 512 x 512 pixels, encompassing a field of view measuring 31° × 31°. The wavelength spectrum spans from 400 to 1000 nm and has a spectral resolution of 204 bands. 

\paragraph{Camera Placement} All cameras are mounted within an aluminum profile construction. Specifically, careful consideration was given to the placement of cameras to maximize overlap between their fields of view and minimize occlusions. Notably, the depth camera is positioned closer to the object to enhance spatial resolution, with objects positioned at least 30cm away from it. 

\paragraph{Illumination} Since the recording with hyperspectral cameras requires illumination from a continuous spectrum, we have integrated 4 halogen (30W) lamps in our setup (yellow lamps in \autoref{fig_Overview_setup}). In addition, LED light strips (light blue) are attached to the aluminium profile construction to offer a homogeneous light source for recording RGB images.

\subsection{Calibration}
\label{Calibration}
Achieving accurate registration of image modalities necessitates the computation of both intrinsic and extrinsic calibration parameters. To facilitate this, we employed a checkerboard target consisting of 16 x 9 checker tiles printed on thick watercolor paper (21 cm x 29.7 cm, 300g/m²). The checkerboard's distinct pattern offers precise visual cues essential for calibration. An example of  recorded calibration images for each camera is depicted in \autoref{fig_Overview_setup} b). Note that the checkerboard pattern is visible in all modalities, which is an important pre-requisite to allow for calibrating each camera to each other camera.

During the calibration process, we recorded 23 different poses of the checkerboard target using all three cameras. Ensuring visibility of the checkerboard in the thermal image posed a unique challenge. To address this, we utilized halogen lamps and initiated the recording process one second after their activation. This deliberate timing allowed the dark tiles of the checkerboard to heat up quickly, rendering them visible in the thermal camera image. 

For the calibration process, we utilized the calibration functions from the open-source library OpenCV utilizing Zhang's algorithm \cite{zhang2000flexible}.
The calibration process yields one intrinsic calibration matrix for each camera and pairwise extrinsic calibration matrices for all cameras. Additionally, we obtain distortion parameters for all cameras. These parameters were used to correct the lens distortions of each  camera. 

\subsection{Dataset}
\label{Dataset}
To evaluate the robustness and general applicability of our registration algorithm across diverse scenarios, we recorded a dataset comprising images of six distinct plant species. This was done to encompass a wide variety of leaf and canopy structures, thus offering a representative sample for evaluation purposes. The recorded dataset can be found on the project github page: \url{https://github.com/eric-stumpe/Plant3DImageReg}. The chosen plant species are as follows:   

\begin{enumerate}
\item Grapevine (\textit{Vitis vinifera})
\item Leopard lily (\textit{Dieffenbachia})
\item Areca palm (\textit{Dypsis lutescens})
\item Philodendron (\textit{Monstera deliciosa})
\item Boston fern (\textit{Nephrolepis}) 
\item Weeping fig (\textit{Ficus benjamina})
\end{enumerate}

Before starting the recording, every plant was placed within the multicamera setup in a way that visibility from each individual camera is ensured. For each recording cycle, first the thermal image was captured to mitigate any heating effects from the halogen lamps. Subsequently, the halogen lamps were turned on to record the hyperspectral image. Following this, infrared and depth images were acquired, followed by the capture of RGB images with the LED lamps illuminated. 

\subsection{Evaluation Metrics}
\label{Evaluation Metrics}

\begin{figure} [ht]
	\centering 
	\includegraphics[width=1\textwidth]{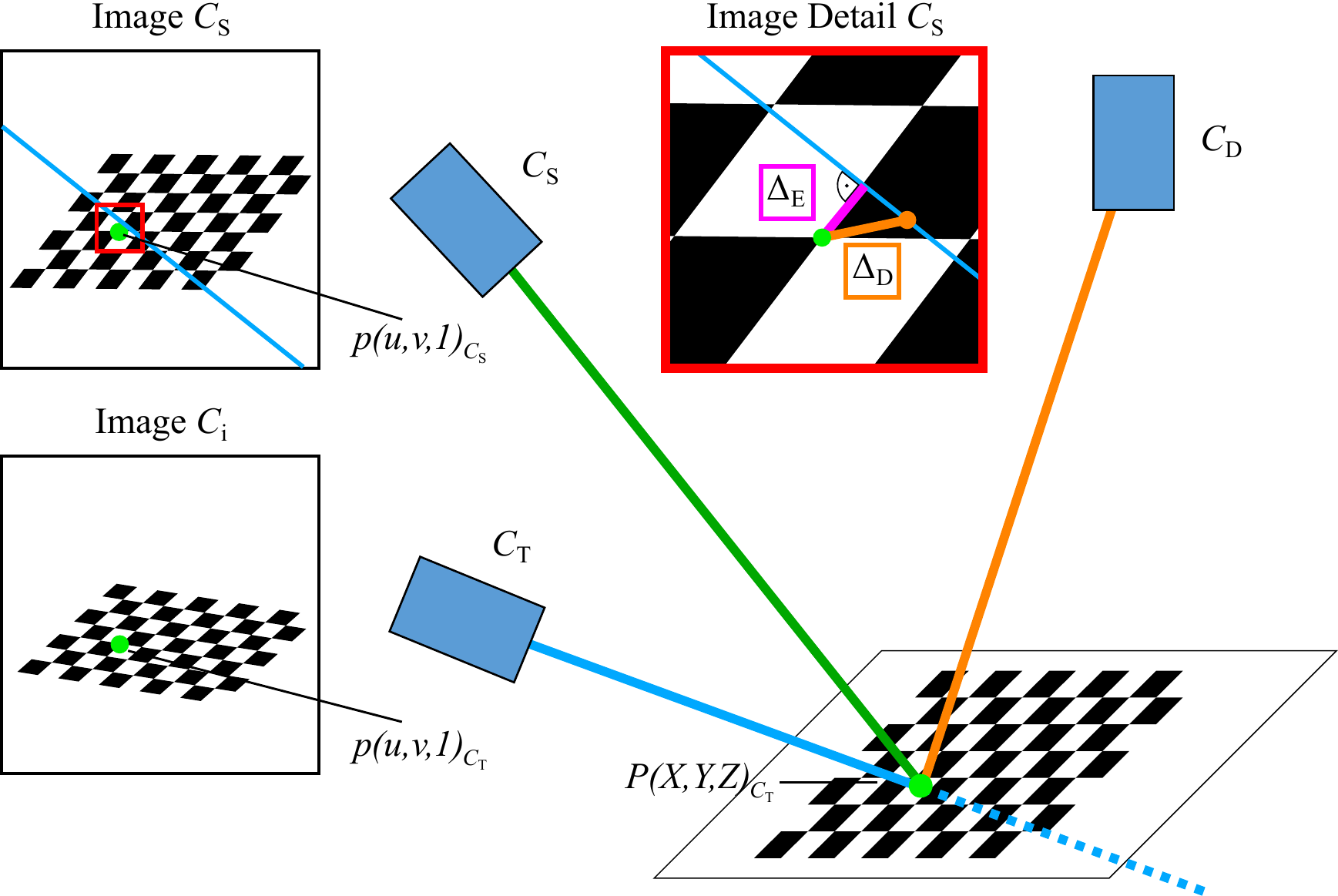}	
	\caption{Visual explanation of the extrinsic and depth errors $\Delta_E$ and $\Delta_D$ for checkerboard calibration patterns. A corner is $p(u,v,1)_{C_T}$ is converted to a 3D epipolar line (blue). This line can be projected into the paired view Image $C_S$.} 
	\label{fig_checker_projection}%
\end{figure}

To assess the accuracy of our developed registration algorithm, we employ an evaluation procedure utilizing all captured  checkerboard images and their corresponding extracted corners. The resulting key metrics are: intrinsic calibration error, extrinsic calibration error, and depth error, see details in the following. Both extrinsic calibration error and depth error are explained with the help of \autoref{fig_checker_projection}. 

\paragraph{Intrinsic Calibration Error}
As mentioned above, we utilize the calibration method integrated from OpenCV, based on Zhang's algorithm \cite{zhang2000flexible}, to determine the intrinsic matrix of each camera. This method sets the 3D world coordinate origin within the known geometry of the placed checkerboard for each image. During calibration, the goal is to minimize the discrepancy between projected and detected corners across all images by finding the optimal intrinsic camera matrix. 
Therefore, the intrinsic calibration error metric we use for evaluation is the 
mean pixel distance between projected and detected corners over all images (a total of 9x6x23 corners)



\paragraph{Extrinsic Calibration Error}
In the extrinsic calibration phase, we estimate the rotation and translation matrices between each possible pair of modalities $C_1$, ...,$C_n$. Since this calibration step does not incorporate  depth information, we compute the error metric based on epipolar lines. For each detected corner in a checkerboard image $C_T$, exemplified by the green point $p(u,v,1)_{C_T}$ in \autoref{fig_checker_projection}, 
we compute the corresponding 3D epipolar line that is visualized as the 3D blue line originating from $C_T$ in \autoref{fig_checker_projection}. This epipolar line is then projected into the other checkerbord image $C_S$ as a 2D line using the computed extrinsic parameters. The error metric is quantified as the minimum distance $\Delta_E$ between the projected epipolar line and the corresponding detected corner in the second image. In \autoref{fig_checker_projection}, this is shown in the ROI (red) of image $C_S$. The shortest distance $\Delta_E$ between the detected point and the epipolar line is perpendicular to that line and is visualized in magenta. The total error is computed by averaging the pixel distances over all corners and images. 

\paragraph{Depth Error}
For each checkerboard we also record the associated depthmap with the Azure Kinect camera. Leveraging our previously described registration methodology, we can directly map pixels between modalities through the utilization of depth information in the form of a 3D mesh. 
Therefore we can take one image $C_T$ and project all corners via ray casting to a second image $C_S$ and and record the pixel distance to the extracted corner. This is visualized in \autoref{fig_checker_projection}, where the process is almost the same as for the extrinsic calibration error. However, here we have now solved for the depth information and can determine the corresponding pixel (orange) in the image detail of $C_S$. The error is then defined by the distance $\Delta_D$ of that pixel to the detected corner. $\Delta_E$ and $\Delta_D$ are closely linked. The more accurate the depth camera the closer $\Delta_D$ will be to $\Delta_E$. Again, the total depth error is computed by averaging pixel distances over all corners and calibration images.

\section{Results}
\label{Results}

In this section we will first present the quantitative results of our computed error metrics from the calibration data. Next, we show different visual results for both images and 3D point clouds from our registration approach in a qualitative manner. Finally, we highlight which parameters influence the overall accuracy of our approach and outline limitations. 

\subsection{Quantitative Results}
\label{Quantitative Results}

Calibration errors are based on pixel measurements. This however does not take into account that images with a higher resolution will naturally yield higher pixel errors. Hence, in addition to the standard evaluation approach with pixel error measurements we also provide normalized results in the brackets of the result tables. The normalized error values are computed by taking the height $h$ and width $w$ of every camera image into account by multiplying every pixel error with $\frac{1000}{\sqrt{h w}}$. The normalized values enable a direct comparison between the scores obtained for different cameras. 

\paragraph{Intrinsic Calibration Error Results}
The intrinsic calibration errors are shown in \autoref{tab:intrinsic} for all four modalities. All error values are in the subpixel range. We get the highest error from the hyperspectral camera. We attribute this finding to an issue which is specific to  the hyperspectral camera, i.e., when taking consecutive recordings of the same scene a shift in the pixels can be observed, which most likely explains the larger error. This shift could be related to the underlying push-broom technology utilized for the Specim IQ camera or other internal mechanic components of the camera.
The error for the thermal camera is slightly lower but still higher than that of infrared and RGB cameras. During calibration we are using halogen lamps to make the checkerboard visible to the thermal camera. Due to the expansion of heat on the checkerboard surface we cannot reach the same sharpness of checker tile corners compared to the cameras that operate in the visible spectrum. When comparing pixel errors with the normalized error (in brackets) we see the largest difference for the RGB image. The normalized error of the RGB camera is the lowest across all modalities, which is due to the RGB camera's high resolution of 3840 x 2160.

\paragraph{Extrinsic Calibration Error Results}
\autoref{tab:extrinsic} displays the extrinsic calibration errors. As outlined in \autoref{Methodology}, we can compute mappings from each modality into every other one. Therefore, we present the errors for every possible combination of camera pairs. Each column denotes the target modality for the mapped corner epipolar lines, while rows indicate the originating modalities. Notably, the lowest errors occur when RGB or infrared modalities are used as the origin (row 1 and 2 respectively), with subpixel accuracy observed for every target modality. Conversely, the highest errors are consistently taking place with mappings involving the hyperspectral image as the target or originating modality (across row 4 and column 4), leading to an increased mean extrinsic error.
Consequently, a large intrinsic error of a camera will also lead to increased extrinsic errors for pairings with the same camera.

\paragraph{Depth Error Results} In \autoref{tab:depth}, the depth error results are presented in the same format as the extrinsic error table. Taking  \autoref{tab:extrinsic} for comparison, we find that the values in all corresponding positions are slightly higher. This is reasonable since the extrinsic error represents the shortest possible distance of each epipolar line to a corresponding corner, whereas the actual projected point could be somewhere else on the epipolar line. Overall, this total difference between both tables is low for mappings between infrared and RGB modalities. For example, from RGB to infrared we get an error of 0.07 pixels for \autoref{tab:extrinsic} and one of 0.16 for \autoref{tab:depth}.
They are also lower for combinations of the thermal and hyperspectral cameras.
Projecting a thermal into the hyperspectral image increases the pixel error from 1.16 for \autoref{tab:extrinsic} to 1.22 in \autoref{tab:depth}. However, for pairings between the two groups of RGB and infrared vs. thermal and hyperspectral camera we see higher differences. Projecting the thermal image into the infrared frame for example we get a pixel error of 0.14 for \autoref{tab:extrinsic} and an error of 0.65 for \autoref{tab:depth}. 
This discrepancy can be attributed to the spatial arrangement of cameras within our multimodal setup. Specifically, RGB and infrared cameras are integrated into the same unit (Azure Kinect), whereas thermal and hyperspectral cameras are positioned farther away from the Kinect camera, but are close to one another. Conceptually, our mapping methodology builds upon the transportation of pixel information via an incoming and outgoing ray. The larger the angle between both rays, the bigger is the influence of inaccurate depth information recorded by the depth sensor leading to larger errors. 

\begin{table}[!htb]

\centering
\caption{Intrinsic Error in pixels and image resolution normalized error in brackets}  \label{tab:intrinsic}
\begin{tabular}{cccc}
Infrared & RGB & Thermal & HS \\
\midrule
0.13 (0.19) & 0.23 (0.08) & 0.22 (0.40) & 0.31 (0.61) \\

\end{tabular}%
\end{table} 

\begin{table}[!htb]
\centering
\caption{Extrinsic Error in pixels and image resolution normalized error in brackets}  \label{tab:extrinsic}
\begin{tabular}{l|cccc}
Origin & Infrared & RGB & Thermal & HS \\
\midrule
Infrared       & -     & 0.22 (0.08)  & 0.23 (0.42)  & 0.26 (0.50)  \\
RGB            & 0.07 (0.01)  & -     & 0.20 (0.36) & 0.46 (0.89)   \\
Thermal        & 0.14 (0.20)  & 0.41 (0.14)  & -     & 1.16 (2.26)   \\
HS             & 0.16 (0.24) & 1.04 (0.36)  & 1.27 (2.29)  & -       \\
\end{tabular}%
\end{table} 

\begin{table}[!htb]
\centering
\caption{Depth Error in pixels and image resolution normalized error in brackets}  \label{tab:depth}
\begin{tabular}{l|cccc}
Origin & Infrared & RGB & Thermal & HS \\
\midrule
Infrared       & -     & 0.55 (0.19)  & 0.93 (1.67) & 0.51 (0.99)   \\
RGB            & 0.16 (0.23)  & -     & 0.96 (1.73)  & 0.64 (1.24)   \\
Thermal        & 0.65 (0.96) & 2.31 (0.80)  & -     & 1.22 (2.38)   \\
HS             & 0.39 (0.58) & 1.60 (0.55) & 1.31 (2.36) & -      \\
\end{tabular}%
\end{table} 

\subsection{Qualitative Results}
\label{Qualitative Results} 

In this section we show visual results of the proposed registration algorithm for our dataset of recorded plants. We present visualizations of registered images, computed masks of error and uncertainty and multimodal point clouds.

\begin{figure}
	\centering 
	\includegraphics[width=1\textwidth]{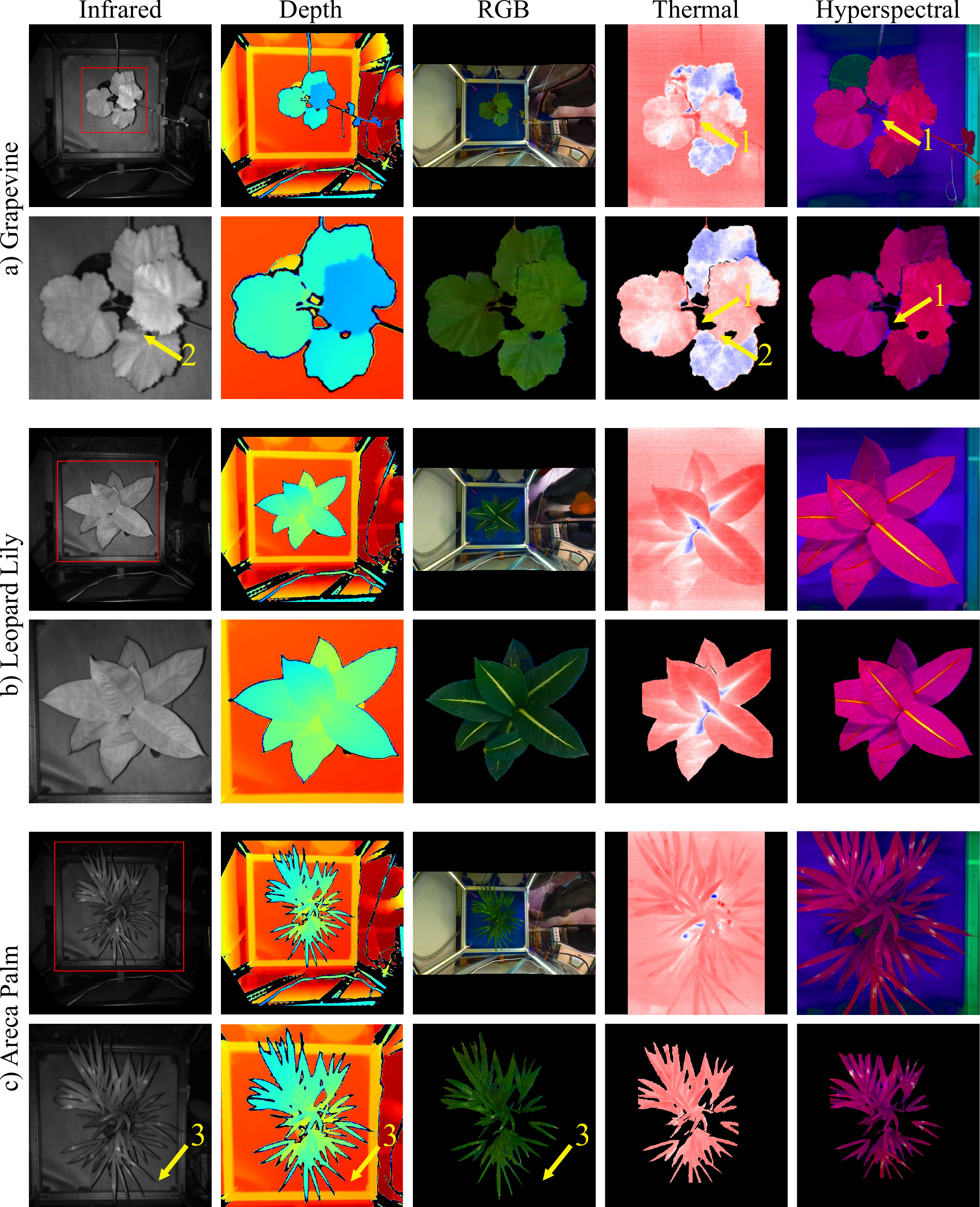}	
	\caption{Image registration results. Per recorded plant species we present the images before (top) and after registration (bottom). Registration is based on the infrared image as the target modality $C_T$. The red boxes define the region of interest that is visualized for the registered images.} 
	\label{fig_image_projections}%
\end{figure}
\begin{figure}
	\centering 
	\includegraphics[width=1\textwidth]{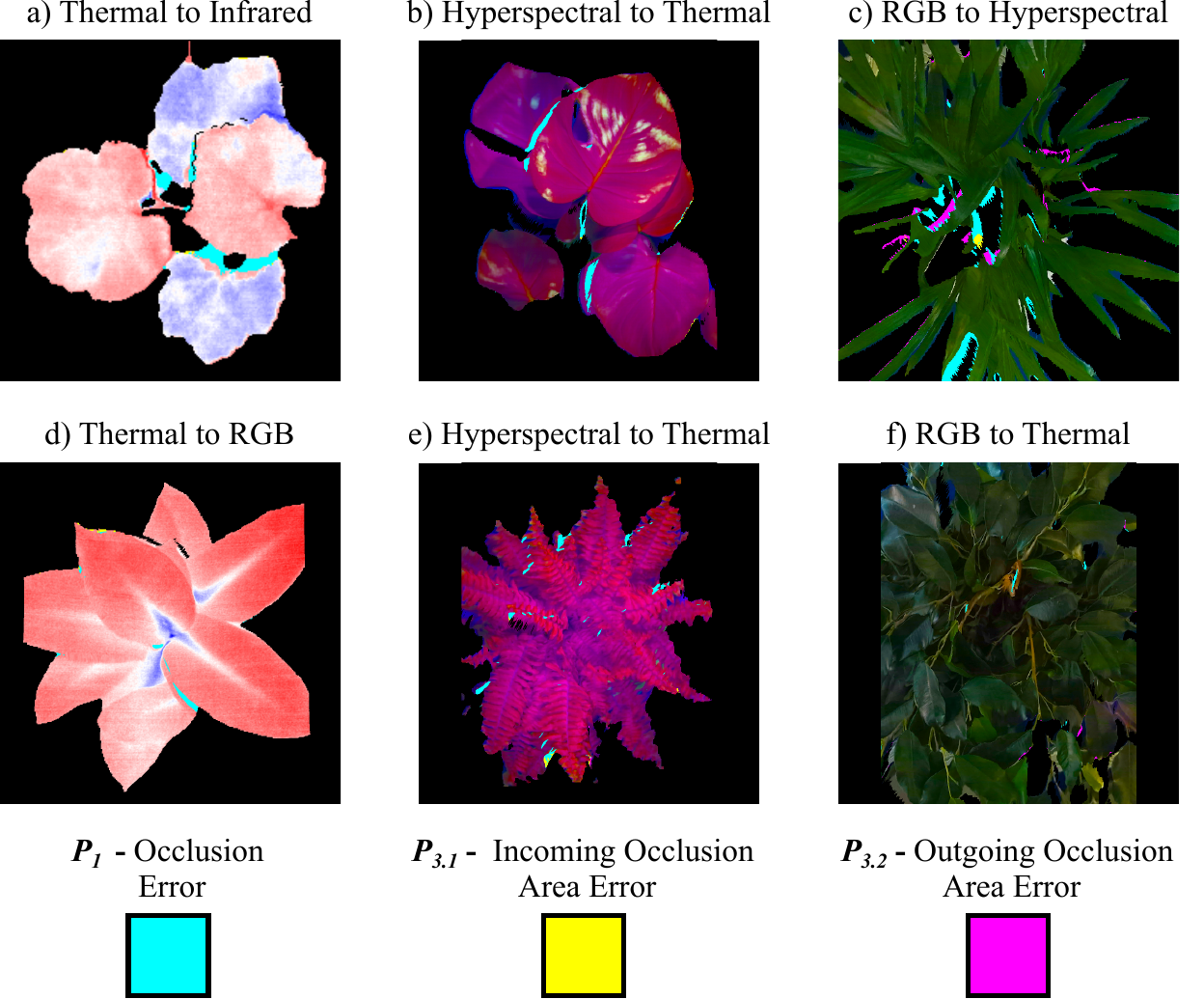}	
	\caption{Detected masks for different projection cases of errors and uncertainties at the example of plants with very different leaf structure.
 For each image the first camera name specifies the source view $C_S$ and the second name the selected target view $C_T$.
 } 
	\label{fig_occlusion_mask_images}%
\end{figure}

\paragraph{Image Registration}
In \autoref{fig_image_projections}, we visualize the mapping results for three distinct plants of our dataset (a: grapevine, b: leopard lily, c: areca palm). The figure illustrates the original input per modality in the top row for each plant, while the bottom row showcases the results of the registration process mapping all modalities to the infrared frame. For better visibility, the registered image view is taken from a selected region of interest (red box). The thermal image is displayed with a diverging colormap that represents increasing temperatures with a color gradient from blue over white to red. Since the hyperspectral image in its original form contains 204 separate channels, we used Principal Component Analysis (PCA) to reduce each image to three channels and present the result as a false color image. During the registration process, only pixels intersecting with the object mesh of the canopy $\mathcal{M}_o$ are mapped, resulting in a black background for pixels without intersections.
The presented images do not yet account for occlusion errors and other projection uncertainties, since we present the results of the mask computation for the different projection cases separately  in \autoref{fig_occlusion_mask_images}. Overall, \autoref{fig_image_projections} shows that the different camera modalities are well-aligned after registration.

In the following we highlight three important aspects that can be observed in this result indicated by markers 1, 2 and 3 (yellow arrows). Marker 1 points to an area in the thermal and hyperspectral image, where we can see the effect of different views leading to a difference in appearance (row 1). Since the hyperspectral camera is positioned further to the left the gap between the two middle leaves appears larger. With conventional image-based transformation methods the registration in these cases will fail, since they do not consider the relationship between 3D objects and the resulting parallax. Our method however can successfully compute a correct registration (row 2).
For marker 2 we can see an example of an occlusion error in the thermal image. Since the bottom (blueish) leaf is occluded by the upper red leaf, pixel information from the upper leaf will be projected back to the infrared frame leading to a spillover of pixel information.
The third marker points at a very thin leaf. The tip of this leaf is visible in the infrared image. However, in depth and RGB image the tip is not visible anymore. When thin structures are present, the ToF camera receives mixed signals from both the object and its background leading to a failure in presenting it correctly. 

\begin{figure}
	\centering 
	\includegraphics[width=1\textwidth]{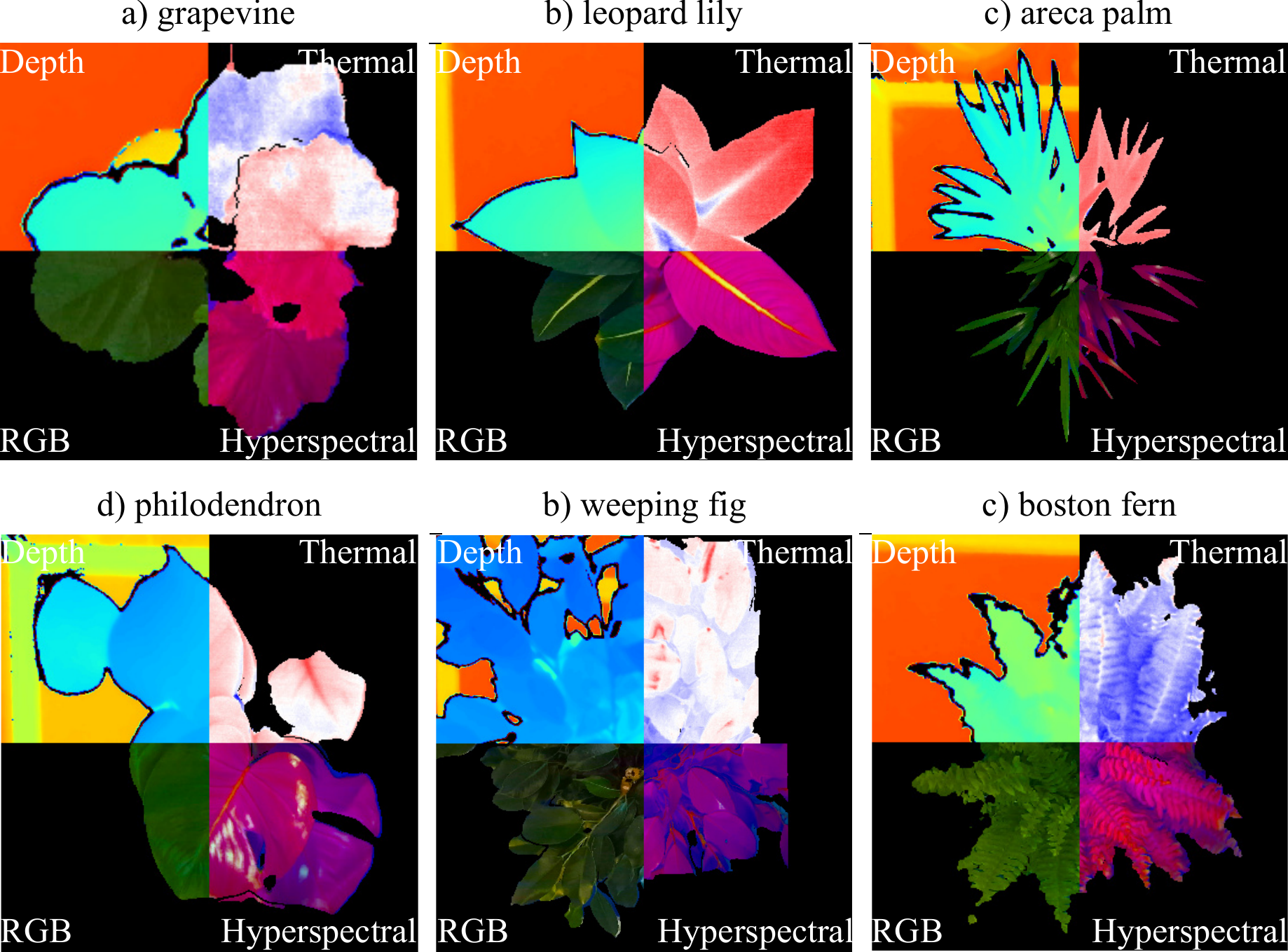}	
	\caption{Image mosaics of four modalities for images registered to the infrared target modality $C_T$. Per plant species an image mosaic of the registered images is constructed to better show the alignment of the modalities and thereby the quality of registration.} 
	\label{fig_image_projections_mosaic}%
\end{figure}

\begin{figure}
	\centering 
	\includegraphics[width=1\textwidth]{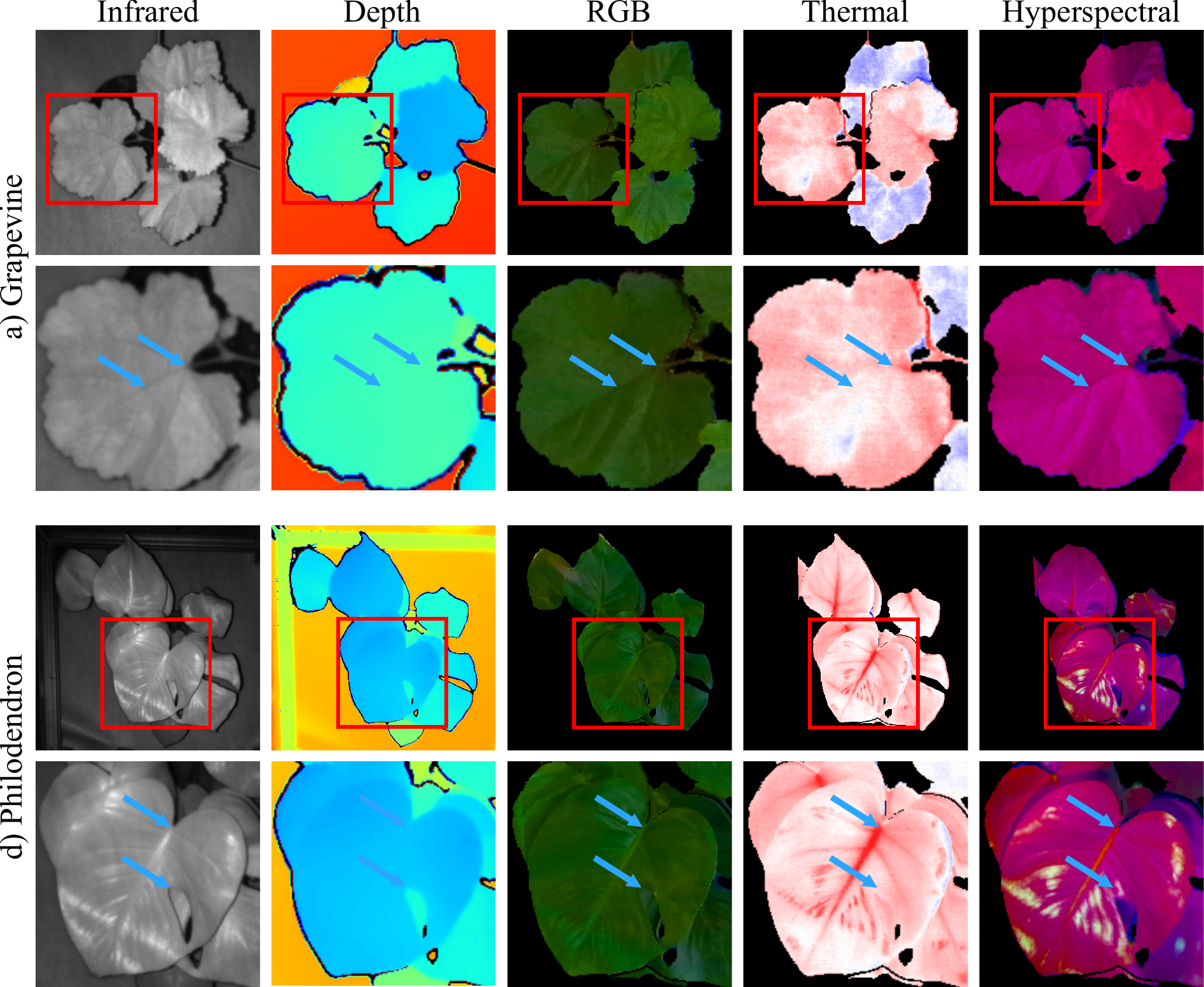}	
	\caption{Detail view of the registration results for all modalities when registered to infrared as the target modality $C_T$. The bottom row for each plant displays a zoomed in view corresponding to the red bounding box. The blue arrows point at the exact same pixel location across modalities. They indicate how well prominent landmarks are pixel aligned after registration.} 
	\label{fig_detail_figure}%
\end{figure}

For better visualization of how well pixels are aligned across modalities, we have constructed image mosaics for the registered depth, thermal, RGB and hyperspectral image in \autoref{fig_image_projections_mosaic} for all 6 plants in our dataset. In the transition regions between mosaic tiles we can see that leaf edges and leaf veins across all images show good alignment. In addition, in \autoref{fig_detail_figure}, we provide a more detailed view of how well pixels are aligned across all modalities. For both grapevine and philodendron we show close up views of the red bounding boxes. The blue arrows are pointing at prominent landmarks such as vein intersection points. The arrows are placed at exactly the same pixel location for each modality. Across all modalities we observe that the indicated landmarks are closely aligned.

\paragraph{Detection of Error and Uncertainty Areas}
In \autoref{fig_occlusion_mask_images}, we present results from our error and uncertainty detection algorithm where occlusions ($P_2$) are highlighted in cyan. Uncertain correspondences in incoming direction ($P_{3.1}$) are depicted in yellow, while those in outgoing direction ($P_{3.2}$) are colored  magenta.
The images show different combinations of source and target modalities.
In \autoref{fig_occlusion_mask_images}a we see that the occluded region that we previously marked with ``1'' in \autoref{fig_image_projections} can be identified and classified correctly.
Plants in \autoref{fig_occlusion_mask_images}b, \autoref{fig_occlusion_mask_images}d and \autoref{fig_occlusion_mask_images}e all have large  leaf structures and thus show occlusions ($P_2$) that can be correctly detected. In such scenarios error types $P_{3.1}$ and $P_{3.2}$ are less common and would only appear if we were to use cameras with a strong difference in position and angle. The areca palm in \autoref{fig_occlusion_mask_images}c on the other hand has many thin overlapping leaves which produced also areas of type $P_{3.1}$ and $P_{3.2}$ which are detected successfully. In \autoref{fig_occlusion_mask_images}f, we observe only a small amount of occluded areas since most of the leaves are positioned at the same height.
\\

\begin{figure}
	\centering 
	\includegraphics[width=1\textwidth]{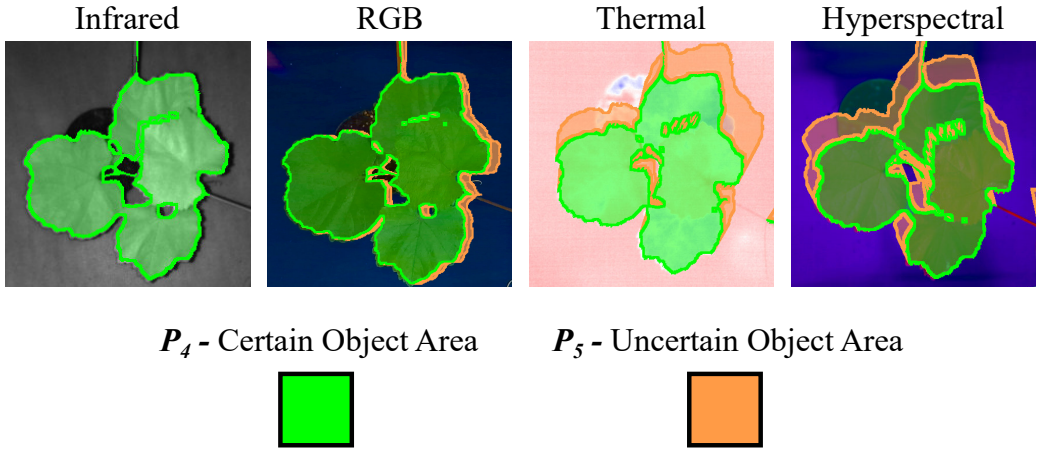}	
	\caption{Visualization of the different Certainty Area types for different target views (Infrared, RGB, Thermal, Hyperspectral). Each image displays color overlays for each projection case: $P_4$: certain object area (green), $P_5$: Uncertain Object Area. $P_6$: Certain Backround Area is represented by the remaining pixels without any overlay.} 
	\label{fig_baseview_regions}%
\end{figure}

In \autoref{fig_baseview_regions}, we visualize projection cases $P_4$: Certain Object Area and $P_5$: Uncertain Object Area, and visualize the respective areas in the image. $P_6$: certain background area is represented by the image parts without any color overlay.
Each image shows the same RoI of the same plant from four different target views in the yet unregistered state. In the infrared image only the certain object area region (green) and the certain background region can be found. This is due to the fact that the infrared modality is already in pixel alignment with the depth image and therefore only displays image regions that are visible to the depth camera. In the RGB image we can see a small Uncertain Object Area region on the right side of the canopy. This area indicates all regions that are not visible to the depth camera and could therefore either belong to the background or to lower level leaves of the canopy. RGB and depth sensor are located within the same camera housing and therefore this region is relatively small. For both thermal and hyperspectral image we find a much larger uncertainty area which can be explained by the larger distance of both these cameras to the depth camera. 

\begin{figure}
	\centering 
	\includegraphics[width=1\textwidth]{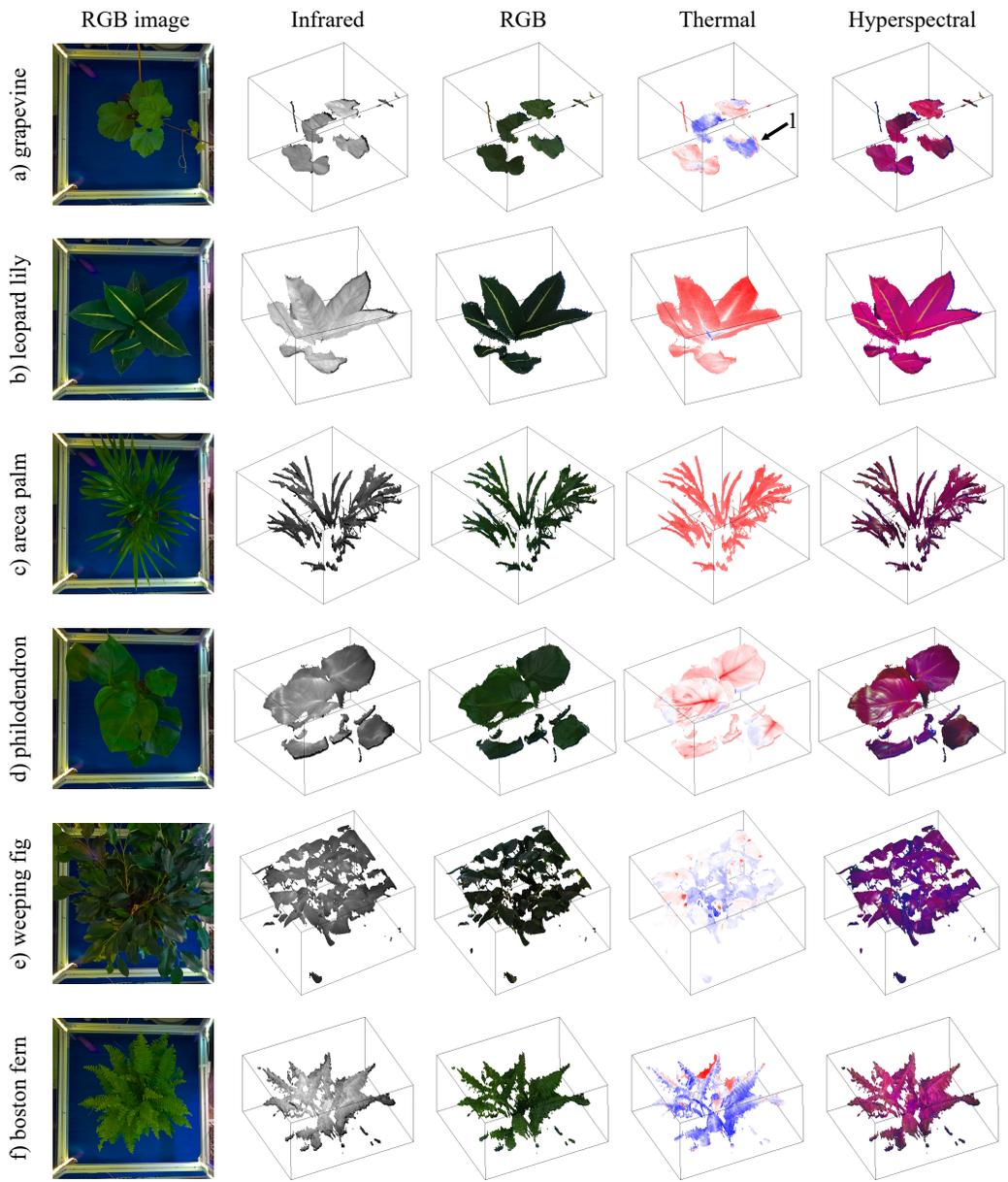}	
	\caption{Registered multimodal point clouds for different plant species. The first columns shows the original RGB images for reference. } 
	\label{fig_3D_representations}%
\end{figure}

\paragraph{Multimodal Point Clouds}
During the registration process for each matched pixel, we automatically compute the intersection of each ray with $\mathcal{M}_o$. By accumulating all intersection points, we can generate a registered 3D point cloud that comprises all modalities. \autoref{fig_3D_representations} displays all modalities (columns) for each registered point cloud for all plant species (rows). 
In the leftmost column we show the RGB image modality for reference. Due to the use of the depth camera we can only generate 3D points for canopy regions that are visible in the depth modality. This explains that some of the displayed leaves are incomplete in the generated 3D point clouds.
In the thermal image in the first row we notice some 3D artifacts in the leaf border regions. For example, the rightmost blue colored leaf has a rim of red colored 3D points (marker 1). These artifacts are related to the utilization of a ToF camera. When the depth map is recorded, the ToF sensor receives mixed signals for pixels that are in-between adjacent leaves at different heights. These artifacts are commonly referred to as "flying pixels" and have been reported before for the Azure Kinect camera in \cite{tolgyessy2021evaluation}. 
 
\subsection{Limitations and Practical Considerations}
\label{Limitations and application considerations} 
In the following paragraphs we report how hardware and different settings are influencing the performance of our registration algorithm. This is done to give the reader some guidelines for the setup of a multimodal camera system.

\paragraph{Depth Camera} Central to our registration method is the use of a depth camera. Consequently, the accuracy of the registration process is highly dependent on the hardware specifications of the depth camera employed. Inaccurate depth information can significantly impact the resulting generated object mesh $\mathcal{M}_o$, leading to errors in the ray intersection point locations that then influence the overall registration accuracy. For instance, the Azure Kinect camera, which we utilized, reports a systematic depth error of $<$11mm + 0.1\% of the distance, which affects the baseline registration accuracy we can achieve. Additionally, the spatial resolution of the depth camera plays a critical role in the overall level of detail achievable in registration, directly influencing the fidelity of the generated object mesh that models the actual canopy surface. Moreover, the choice of depth camera technology influences the accuracy of registering leaf border regions, as artifacts like "flying pixel" errors may occur in these areas. 

\paragraph{Calibration} The calibration process plays a crucial role in ensuring a high accuracy of image registration. It is important that the calibration board used should be as planar as possible. In addition, the corner detection algorithm employed should reliably identify corners even in uncommon modalities such as thermal images. Furthermore, during calibration, the positioning of the calibration board must encompass the entire  depth range expected during recording. Incompletely capturing the depth range may result in lens and distortion parameters that do not accurately enforce correct projections for the actual plant recordings. Moreover, any displacement and motion of cameras post-calibration must be avoided to maintain the validity of the computed extrinsic parameters. 

\paragraph{Pixel Position Consistency} Observations with the hyperspectral camera revealed that between consecutive recordings from the exact same position, a shift in image content by multiple pixels can occur. This phenomenon is suspected to be caused by the internal push-broom line scanner technology. We observed a similar problem in the infrared and depth image with the Intel Realsense L515 depth camera, which we used in experiments before integrating the Azure Kinect camera. Such shifts directly impact the accuracy of calibration and registration processes since they cause inconsistent corner positions and leaf pixels in the registered image. 

\paragraph{Recording Time} Due to the nonrigid nature of plant leaves, they are susceptible to movement induced by external factors like airflow. Hence, it is crucial to minimize the time intervals between recordings of each camera in the setup to mitigate registration offsets. 

\paragraph{Illumination and Surface Specularity}
Illumination in a multimodal camera setup also plays a significant role. In our setup we are using halogen lamps to illuminate the scene for the hyperspectral camera. However, the halogen lamps radiate significant heat. Hence it was necessary to do the thermal recording before hyperspectral imaging to measure the canopy temperature without any extra influence. Compared to materials such as polished metals, plant leaves are characterized by a mostly diffuse surface.
With many depth technologies including ToF, 3D information cannot be properly measured in highly specular surfaces.

These considerations highlight the importance of considering all technical challenges and limitations to ensure the accuracy and reliability of the registration process in plant imaging applications. 

\section{Conclusion}
\label{Conclusion}

In this work we have presented a novel algorithm to register images from different camera modalities for plant phenotyping. Unlike most previous registration algorithms, our method is not reliant on detecting specific patterns in each modality (like feature points) and can therefore be utilized for any type of recorded plant or image modality. We achieve this by utilizing a depth camera to create a 3D object mesh from the plant canopy and mapping pixels via ray casting. This 3D approach further enables us to identify and filter out occluded regions and other areas of uncertainty, which is not possible with 2D registration methods.  To this end, we have introduced a classification of different possible special cases that can occur during 3D projection and that may lead to occlusions or uncertainties in the registration. 
Moreover, our method facilitates the computation of both registered images and registered point clouds, enhancing the utility of the multi-camera and multimodal image data. The only prerequisite for using the proposed algorithm is to establish a hardware setting where all cameras are in a fixed position with overlapping viewport and to calibrate all cameras (intrinsically and extrinsically) before starting the recording process. 
We have evaluated our approach both quantitatively (based on calibration data) and qualitatively by using a diverse set of plant species with heterogeneous plant structure and shape. Depending on which camera combination is used, the overall registration error we obtained in our experiments (depth error) ranges between 0.16 and 2.31 pixels. Visually, the registered images visually also show a strong agreement between the individual camera modalities.

A future challenge we aim to solve is the occurrence of 3D artifacts introduced by the depth camera, such as “flying pixels” that occur in leaf border regions. To this end, we propose to develop a leaf segmentation algorithm that enables us to easily identify leaf border regions in images to mask out such artifacts. 

Overall, we envision that our registration method will serve the generation of multimodal plant image datasets. Such datasets will enable pixel-accurate cross-modal comparisons and are urgently needed to design and develop multimodal approaches for e.g., leaf segmentation, leaf tracking, and plant stress/disease classification algorithms.


\section*{Acknowledgements}
This research was funded in whole by the Research Promotion Agency of Lower Austria (GFF) through project ``\textit{DiPhSpec -  Digitizing physiology for advanced spectral plant diagnostics}'' (project no. FTI18-005) and PhD scholarship ``\textit{Visual Monitoring of Crop Production Systems: A Multimodal Approach}''  (project no. FTI21-D-021).

\appendix



\bibliographystyle{elsarticle-harv} 
\bibliography{bibliography}

\begin{thebibliography}{27}
\expandafter\ifx\csname natexlab\endcsname\relax\def\natexlab#1{#1}\fi
\providecommand{\url}[1]{\texttt{#1}}
\providecommand{\href}[2]{#2}
\providecommand{\path}[1]{#1}
\providecommand{\DOIprefix}{doi:}
\providecommand{\ArXivprefix}{arXiv:}
\providecommand{\URLprefix}{URL: }
\providecommand{\Pubmedprefix}{pmid:}
\providecommand{\doi}[1]{\href{http://dx.doi.org/#1}{\path{#1}}}
\providecommand{\Pubmed}[1]{\href{pmid:#1}{\path{#1}}}
\providecommand{\bibinfo}[2]{#2}
\ifx\xfnm\relax \def\xfnm[#1]{\unskip,\space#1}\fi
\bibitem[{Azam et~al.(2019)Azam, Munir, Sheri, Ko, Hussain and Jeon}]{azam2019data}
\bibinfo{author}{Azam, S.}, \bibinfo{author}{Munir, F.}, \bibinfo{author}{Sheri, A.M.}, \bibinfo{author}{Ko, Y.}, \bibinfo{author}{Hussain, I.}, \bibinfo{author}{Jeon, M.}, \bibinfo{year}{2019}.
\newblock \bibinfo{title}{Data fusion of lidar and thermal camera for autonomous driving}, in: \bibinfo{booktitle}{Applied Industrial Optics: Spectroscopy, Imaging and Metrology}, \bibinfo{organization}{Optica Publishing Group}. pp. \bibinfo{pages}{T2A--5}.
\bibitem[{Behmann et~al.(2016)Behmann, Mahlein, Paulus, Dupuis, Kuhlmann, Oerke and Pl{\"u}mer}]{behmann2016generation}
\bibinfo{author}{Behmann, J.}, \bibinfo{author}{Mahlein, A.K.}, \bibinfo{author}{Paulus, S.}, \bibinfo{author}{Dupuis, J.}, \bibinfo{author}{Kuhlmann, H.}, \bibinfo{author}{Oerke, E.C.}, \bibinfo{author}{Pl{\"u}mer, L.}, \bibinfo{year}{2016}.
\newblock \bibinfo{title}{Generation and application of hyperspectral 3d plant models: methods and challenges}.
\newblock \bibinfo{journal}{Machine Vision and Applications} \bibinfo{volume}{27}, \bibinfo{pages}{611--624}.
\bibitem[{Van~den Bergh and Van~Gool(2011)}]{van2011combining}
\bibinfo{author}{Van~den Bergh, M.}, \bibinfo{author}{Van~Gool, L.}, \bibinfo{year}{2011}.
\newblock \bibinfo{title}{Combining rgb and tof cameras for real-time 3d hand gesture interaction}, in: \bibinfo{booktitle}{2011 IEEE workshop on applications of computer vision (WACV)}, \bibinfo{organization}{IEEE}. pp. \bibinfo{pages}{66--72}.
\bibitem[{Borrmann et~al.(2016)Borrmann, Leutert, Schilling and N{\"u}chter}]{borrmann2016spatial}
\bibinfo{author}{Borrmann, D.}, \bibinfo{author}{Leutert, F.}, \bibinfo{author}{Schilling, K.}, \bibinfo{author}{N{\"u}chter, A.}, \bibinfo{year}{2016}.
\newblock \bibinfo{title}{Spatial projection of thermal data for visual inspection}, in: \bibinfo{booktitle}{2016 14th International Conference on Control, Automation, Robotics and Vision (ICARCV)}, \bibinfo{organization}{IEEE}. pp. \bibinfo{pages}{1--6}.
\bibitem[{Clamens et~al.(2021)Clamens, Alexakis, Duverne, Seulin, Fauvet and Fofi}]{clamens2021real}
\bibinfo{author}{Clamens, T.}, \bibinfo{author}{Alexakis, G.}, \bibinfo{author}{Duverne, R.}, \bibinfo{author}{Seulin, R.}, \bibinfo{author}{Fauvet, E.}, \bibinfo{author}{Fofi, D.}, \bibinfo{year}{2021}.
\newblock \bibinfo{title}{Real-time multispectral image processing and registration on 3d point cloud for vineyard analysis.}, in: \bibinfo{booktitle}{VISIGRAPP (4: VISAPP)}, pp. \bibinfo{pages}{388--398}.
\bibitem[{Cucho-Padin et~al.(2020)Cucho-Padin, Rinza, Ninanya, Loayza, Quiroz and Ram{\'\i}rez}]{cucho2020development}
\bibinfo{author}{Cucho-Padin, G.}, \bibinfo{author}{Rinza, J.}, \bibinfo{author}{Ninanya, J.}, \bibinfo{author}{Loayza, H.}, \bibinfo{author}{Quiroz, R.}, \bibinfo{author}{Ram{\'\i}rez, D.A.}, \bibinfo{year}{2020}.
\newblock \bibinfo{title}{Development of an open-source thermal image processing software for improving irrigation management in potato crops (solanum tuberosum l.)}.
\newblock \bibinfo{journal}{Sensors} \bibinfo{volume}{20}, \bibinfo{pages}{472}.
\bibitem[{Dandrifosse et~al.(2021)Dandrifosse, Carlier, Dumont and Mercatoris}]{dandrifosse2021registration}
\bibinfo{author}{Dandrifosse, S.}, \bibinfo{author}{Carlier, A.}, \bibinfo{author}{Dumont, B.}, \bibinfo{author}{Mercatoris, B.}, \bibinfo{year}{2021}.
\newblock \bibinfo{title}{Registration and fusion of close-range multimodal wheat images in field conditions}.
\newblock \bibinfo{journal}{Remote Sensing} \bibinfo{volume}{13}, \bibinfo{pages}{1380}.
\bibitem[{Fedorov et~al.(2003)Fedorov, Fonseca, Kenney and Manjunath}]{fedorov2003automatic}
\bibinfo{author}{Fedorov, D.V.}, \bibinfo{author}{Fonseca, L.M.G.}, \bibinfo{author}{Kenney, C.}, \bibinfo{author}{Manjunath, B.S.}, \bibinfo{year}{2003}.
\newblock \bibinfo{title}{Automatic registration and mosaicking system for remotely sensed imagery}, in: \bibinfo{booktitle}{Image and Signal Processing for Remote Sensing VIII}, \bibinfo{organization}{SPIE}. pp. \bibinfo{pages}{444--451}.
\bibitem[{Gan et~al.(2018)Gan, Lee and Alchanatis}]{gan2018photogrammetry}
\bibinfo{author}{Gan, H.}, \bibinfo{author}{Lee, W.S.}, \bibinfo{author}{Alchanatis, V.}, \bibinfo{year}{2018}.
\newblock \bibinfo{title}{A photogrammetry-based image registration method for multi-camera systems--with applications in images of a tree crop}.
\newblock \bibinfo{journal}{Biosystems engineering} \bibinfo{volume}{174}, \bibinfo{pages}{89--106}.
\bibitem[{Kuglin(1975)}]{kuglin1975phase}
\bibinfo{author}{Kuglin, C.D.}, \bibinfo{year}{1975}.
\newblock \bibinfo{title}{The phase correlation image alignment method}, in: \bibinfo{booktitle}{IEEE Int. Conf. on Cybernetics and Society, 1975}, pp. \bibinfo{pages}{163--165}.
\bibitem[{Kyriacou et~al.(1999)Kyriacou, Davatzikos, Zinreich and Bryan}]{kyriacou1999nonlinear}
\bibinfo{author}{Kyriacou, S.K.}, \bibinfo{author}{Davatzikos, C.}, \bibinfo{author}{Zinreich, S.J.}, \bibinfo{author}{Bryan, R.N.}, \bibinfo{year}{1999}.
\newblock \bibinfo{title}{Nonlinear elastic registration of brain images with tumor pathology using a biomechanical model [mri]}.
\newblock \bibinfo{journal}{IEEE transactions on medical imaging} \bibinfo{volume}{18}, \bibinfo{pages}{580--592}.
\bibitem[{Lin et~al.(2019)Lin, Jarzabek-Rychard, Tong and Maas}]{lin2019fusion}
\bibinfo{author}{Lin, D.}, \bibinfo{author}{Jarzabek-Rychard, M.}, \bibinfo{author}{Tong, X.}, \bibinfo{author}{Maas, H.G.}, \bibinfo{year}{2019}.
\newblock \bibinfo{title}{Fusion of thermal imagery with point clouds for building fa{\c{c}}ade thermal attribute mapping}.
\newblock \bibinfo{journal}{ISPRS journal of photogrammetry and remote sensing} \bibinfo{volume}{151}, \bibinfo{pages}{162--175}.
\bibitem[{Liu et~al.(2023)Liu, Feng, Sun, Li, Ru and Xu}]{liu2023yolactfusion}
\bibinfo{author}{Liu, C.}, \bibinfo{author}{Feng, Q.}, \bibinfo{author}{Sun, Y.}, \bibinfo{author}{Li, Y.}, \bibinfo{author}{Ru, M.}, \bibinfo{author}{Xu, L.}, \bibinfo{year}{2023}.
\newblock \bibinfo{title}{Yolactfusion: An instance segmentation method for rgb-nir multimodal image fusion based on an attention mechanism}.
\newblock \bibinfo{journal}{Computers and Electronics in Agriculture} \bibinfo{volume}{213}, \bibinfo{pages}{108186}.
\bibitem[{Liu et~al.(2018)Liu, Lee and Chahl}]{liu2018registration}
\bibinfo{author}{Liu, H.}, \bibinfo{author}{Lee, S.H.}, \bibinfo{author}{Chahl, J.S.}, \bibinfo{year}{2018}.
\newblock \bibinfo{title}{Registration of multispectral 3d points for plant inspection}.
\newblock \bibinfo{journal}{Precision agriculture} \bibinfo{volume}{19}, \bibinfo{pages}{513--536}.
\bibitem[{Maes et~al.(1997)Maes, Collignon, Vandermeulen, Marchal and Suetens}]{maes1997multimodality}
\bibinfo{author}{Maes, F.}, \bibinfo{author}{Collignon, A.}, \bibinfo{author}{Vandermeulen, D.}, \bibinfo{author}{Marchal, G.}, \bibinfo{author}{Suetens, P.}, \bibinfo{year}{1997}.
\newblock \bibinfo{title}{Multimodality image registration by maximization of mutual information}.
\newblock \bibinfo{journal}{IEEE transactions on Medical Imaging} \bibinfo{volume}{16}, \bibinfo{pages}{187--198}.
\bibitem[{Maimaitijiang et~al.(2020)Maimaitijiang, Sagan, Sidike, Hartling, Esposito and Fritschi}]{maimaitijiang2020soybean}
\bibinfo{author}{Maimaitijiang, M.}, \bibinfo{author}{Sagan, V.}, \bibinfo{author}{Sidike, P.}, \bibinfo{author}{Hartling, S.}, \bibinfo{author}{Esposito, F.}, \bibinfo{author}{Fritschi, F.B.}, \bibinfo{year}{2020}.
\newblock \bibinfo{title}{Soybean yield prediction from uav using multimodal data fusion and deep learning}.
\newblock \bibinfo{journal}{Remote sensing of environment} \bibinfo{volume}{237}, \bibinfo{pages}{111599}.
\bibitem[{Qiu et~al.(2021)Qiu, Miao, Zhang and Li}]{qiu2021detection}
\bibinfo{author}{Qiu, R.}, \bibinfo{author}{Miao, Y.}, \bibinfo{author}{Zhang, M.}, \bibinfo{author}{Li, H.}, \bibinfo{year}{2021}.
\newblock \bibinfo{title}{Detection of the 3d temperature characteristics of maize under water stress using thermal and rgb-d cameras}.
\newblock \bibinfo{journal}{Computers and Electronics in Agriculture} \bibinfo{volume}{191}, \bibinfo{pages}{106551}.
\bibitem[{Salve et~al.(2022)Salve, Yannawar and Sardesai}]{salve2022multimodal}
\bibinfo{author}{Salve, P.}, \bibinfo{author}{Yannawar, P.}, \bibinfo{author}{Sardesai, M.}, \bibinfo{year}{2022}.
\newblock \bibinfo{title}{Multimodal plant recognition through hybrid feature fusion technique using imaging and non-imaging hyper-spectral data}.
\newblock \bibinfo{journal}{Journal of King Saud University-Computer and Information Sciences} \bibinfo{volume}{34}, \bibinfo{pages}{1361--1369}.
\bibitem[{Sch{\"o}nauer et~al.(2013)Sch{\"o}nauer, Vonach, Gerstweiler and Kaufmann}]{schonauer20133d}
\bibinfo{author}{Sch{\"o}nauer, C.}, \bibinfo{author}{Vonach, E.}, \bibinfo{author}{Gerstweiler, G.}, \bibinfo{author}{Kaufmann, H.}, \bibinfo{year}{2013}.
\newblock \bibinfo{title}{3d building reconstruction and thermal mapping in fire brigade operations}, in: \bibinfo{booktitle}{Proceedings of the 4th Augmented Human International Conference}, pp. \bibinfo{pages}{202--205}.
\bibitem[{Sharma et~al.(2023)Sharma, Banerjee, Hayden and Kant}]{sharma2023open}
\bibinfo{author}{Sharma, N.}, \bibinfo{author}{Banerjee, B.P.}, \bibinfo{author}{Hayden, M.}, \bibinfo{author}{Kant, S.}, \bibinfo{year}{2023}.
\newblock \bibinfo{title}{An open-source package for thermal and multispectral image analysis for plants in glasshouse}.
\newblock \bibinfo{journal}{Plants} \bibinfo{volume}{12}, \bibinfo{pages}{317}.
\bibitem[{T{\"o}lgyessy et~al.(2021)T{\"o}lgyessy, Dekan, Chovanec and Hubinsk{\`y}}]{tolgyessy2021evaluation}
\bibinfo{author}{T{\"o}lgyessy, M.}, \bibinfo{author}{Dekan, M.}, \bibinfo{author}{Chovanec, L.}, \bibinfo{author}{Hubinsk{\`y}, P.}, \bibinfo{year}{2021}.
\newblock \bibinfo{title}{Evaluation of the azure kinect and its comparison to kinect v1 and kinect v2}.
\newblock \bibinfo{journal}{Sensors} \bibinfo{volume}{21}, \bibinfo{pages}{413}.
\bibitem[{Vuleti{\'c} et~al.(2023)Vuleti{\'c}, Car and Orsag}]{vuletic2023close}
\bibinfo{author}{Vuleti{\'c}, J.}, \bibinfo{author}{Car, M.}, \bibinfo{author}{Orsag, M.}, \bibinfo{year}{2023}.
\newblock \bibinfo{title}{Close-range multispectral imaging with multispectral-depth (ms-d) system}.
\newblock \bibinfo{journal}{Biosystems Engineering} \bibinfo{volume}{231}, \bibinfo{pages}{178--194}.
\bibitem[{Wang et~al.(2023)Wang, Miao, Han, Li, Zhang and Peng}]{wang2023extraction}
\bibinfo{author}{Wang, L.}, \bibinfo{author}{Miao, Y.}, \bibinfo{author}{Han, Y.}, \bibinfo{author}{Li, H.}, \bibinfo{author}{Zhang, M.}, \bibinfo{author}{Peng, C.}, \bibinfo{year}{2023}.
\newblock \bibinfo{title}{Extraction of 3d distribution of potato plant cwsi based on thermal infrared image and binocular stereovision system}.
\newblock \bibinfo{journal}{Frontiers in Plant Science} \bibinfo{volume}{13}, \bibinfo{pages}{1104390}.
\bibitem[{Xie et~al.(2023)Xie, Du, Ma and Cen}]{xie2023generating}
\bibinfo{author}{Xie, P.}, \bibinfo{author}{Du, R.}, \bibinfo{author}{Ma, Z.}, \bibinfo{author}{Cen, H.}, \bibinfo{year}{2023}.
\newblock \bibinfo{title}{Generating 3d multispectral point clouds of plants with fusion of snapshot spectral and rgb-d images}.
\newblock \bibinfo{journal}{Plant Phenomics} \bibinfo{volume}{5}, \bibinfo{pages}{0040}.
\bibitem[{Yang et~al.(2009)Yang, Wang, Wheaton, Cooley and Moran}]{yang2009automatic}
\bibinfo{author}{Yang, W.}, \bibinfo{author}{Wang, X.}, \bibinfo{author}{Wheaton, A.}, \bibinfo{author}{Cooley, N.}, \bibinfo{author}{Moran, B.}, \bibinfo{year}{2009}.
\newblock \bibinfo{title}{Automatic optical and ir image fusion for plant water stress analysis}, in: \bibinfo{booktitle}{2009 12th International Conference on Information Fusion}, \bibinfo{organization}{IEEE}. pp. \bibinfo{pages}{1053--1059}.
\bibitem[{Zhang(2000)}]{zhang2000flexible}
\bibinfo{author}{Zhang, Z.}, \bibinfo{year}{2000}.
\newblock \bibinfo{title}{A flexible new technique for camera calibration}.
\newblock \bibinfo{journal}{IEEE Transactions on pattern analysis and machine intelligence} \bibinfo{volume}{22}, \bibinfo{pages}{1330--1334}.
\bibitem[{Zitova and Flusser(2003)}]{zitova2003image}
\bibinfo{author}{Zitova, B.}, \bibinfo{author}{Flusser, J.}, \bibinfo{year}{2003}.
\newblock \bibinfo{title}{Image registration methods: a survey}.
\newblock \bibinfo{journal}{Image and vision computing} \bibinfo{volume}{21}, \bibinfo{pages}{977--1000}.

\end{thebibliography}






\end{document}